\documentclass[lettersize,journal]{IEEEtran}
\usepackage{cite}
\usepackage{amsmath,amssymb,amsfonts}
\usepackage{algorithmic}
\usepackage{graphicx}
\usepackage{textcomp}
\usepackage{booktabs}

\usepackage{amssymb}
\usepackage{amsmath}
\usepackage[table]{xcolor}
\usepackage{multirow}
\usepackage{pifont}
\usepackage{wrapfig}

\newcommand{\eg}{{\em e.g.}}           
\newcommand{\ie}{{\em i.e.}}           

\begin{document}
\title{Diversity-enhanced Collaborative Mamba for Semi-supervised Medical Image Segmentation}
\author{Shumeng Li, Jian Zhang, Lei Qi, Luping Zhou, Yinghuan Shi$^*$, Yang Gao
\thanks{This work was supported by NSFC Project (62222604, 62206052), China Postdoctoral Science Foundation (2024M750424), Fundamental Research Funds for the Central Universities (020214380120, 020214380128), State Key Laboratory Fund (ZZKT2024A14, ZZKT2025B05), Postdoctoral Fellowship Program of CPSF (GZC20240252), Jiangsu Funding Program for Excellent Postdoctoral Talent (2024ZB242), and Jiangsu Science and Technology Project (BF2025061, BG2024031).}
\thanks{Shumeng Li, Jian Zhang, Yinghuan Shi, and Yang Gao are with the State Key Laboratory of Novel Software Technology, and National Institute of Healthcare Data Science, Nanjing University, China. Jian Zhang and Yang Gao are also with the School of Intelligence Science and Technology, Nanjing University, China. (E-mail: lism@smail.nju.edu.cn, zhang.jian@nju.edu.cn, syh@nju.edu.cn, gaoy@nju.edu.cn)}
\thanks{Lei Qi is with the School of Computer Science and Engineering, and the Key Lab of Computer Network and Information Integration (Ministry of Education), Southeast University, China. (E-mail: qilei@seu.edu.cn)}
\thanks{Luping Zhou is with the School of Electrical and Information Engineering, The University of Sydney, Australia. (E-mail: luping.zhou@sydney.edu.au)}
\thanks{The corresponding author of this work is Yinghuan Shi.}
}

\maketitle

\begin{abstract}
Acquiring high-quality annotated data for medical image segmentation is tedious and costly. Semi-supervised segmentation techniques alleviate this burden by leveraging unlabeled data to generate pseudo labels.
Recently, advanced state space models, represented by Mamba, have shown efficient handling of long-range dependencies. This drives us to explore their potential in semi-supervised medical image segmentation.
In this paper, we propose a novel Diversity-enhanced Collaborative Mamba framework (namely DCMamba) for semi-supervised medical image segmentation, which explores and utilizes the diversity from data, network, and feature perspectives. 
Firstly, from the data perspective, we develop patch-level weak-strong mixing augmentation with Mamba's scanning modeling characteristics. 
Moreover, from the network perspective, we introduce a diverse-scan collaboration module, which could benefit from the prediction discrepancies arising from different scanning directions. 
Furthermore, from the feature perspective, we adopt an uncertainty-weighted contrastive learning mechanism to enhance the diversity of feature representation. 
Experiments demonstrate that our DCMamba significantly outperforms other semi-supervised medical image segmentation methods, e.g., yielding the latest SSM-based method by 6.69\% on the Synapse dataset with 20\% labeled data. 
The code is available at {\color{magenta}https://github.com/ShumengLI/DCMamba}. 
\end{abstract}


\section{Introduction}
\label{sec:introduction}
\IEEEPARstart{R}{ecently}, State space model (SSMs)~\cite{mehta2023long, wang2023selective, gu2023mamba, liu2024vmamba}, represented by Mamba~\cite{gu2023mamba} and VMamba~\cite{liu2024vmamba}, have emerged as a promising approach for medical image segmentation, owing to their efficient long-range modeling capabilities. SSM-based methods~\cite{liu2024vmamba, ruan2024vm, wang2024mamba, hu2024zigma, pei2024efficientvmamba} exhibit linear computational complexity as the context length increases, alleviating the limitations of the local receptive fields of convolution neural networks (CNNs) and the high computational cost of self-attention in Transformers.
The success of visual SSM-based methods like VMamba~\cite{liu2024vmamba} benefits from the 2D Selective Scan (SS2D) module, which enhances the global modeling ability through a four-way scanning mechanism. 
With SS2D as a core component, SSM-based methods perform competitively in medical image segmentation (\eg, SSM-based U-Net~\cite{ruan2024vm, wang2024mamba}). 

\begin{figure}[t]
   \centering
   \includegraphics[width=1.0\linewidth]{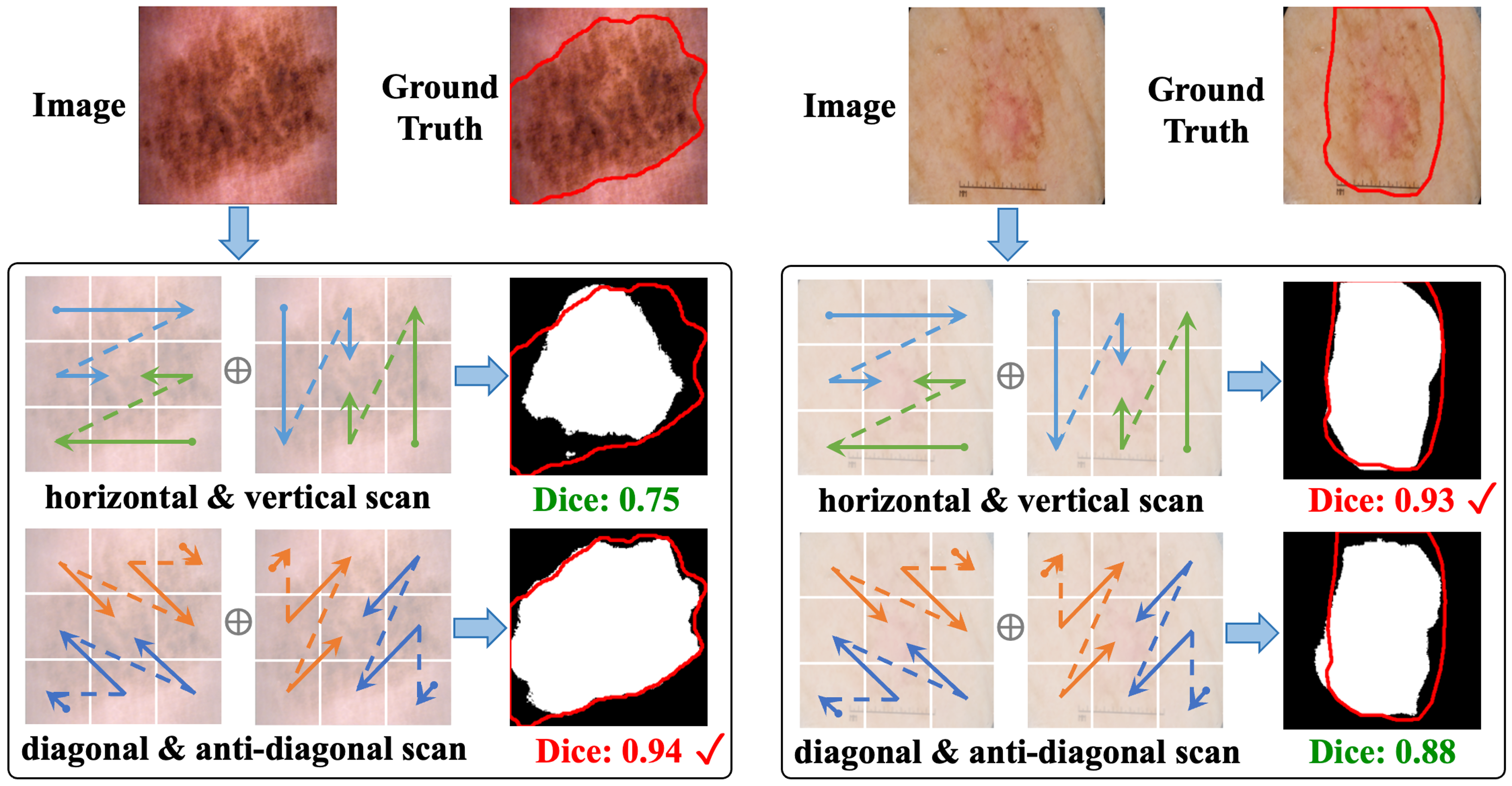}
   \caption{Analysis of different scanning directions of SS2D module for medical images. We predict the same sample using two independently trained networks with different scanning directions. \textbf{Left}: The network with diagonal and anti-diagonal scanning achieves a better performance than horizontal and vertical scanning. \textbf{Right}: The network with horizontal and vertical scanning outperforms diagonal and anti-diagonal scanning. It can be observed that different scanning directions and visual appearances of objects are correlated, which suggests the potential benefits of incorporating multiple scanning directions. }
   \label{fig: intro}
\end{figure}

These methods have made remarkable progress in fully-supervised medical image segmentation~\cite{ma2024u, ruan2024vm, wang2024mamba, yang2024cardiovascular, liu2024swin}, but obtaining a large amount of high-quality annotated data has always been a significant challenge~\cite{jiao2023learning}. 
Semi-supervised learning (SSL) effectively leverages unlabeled data with the guidance of limited labeled data to reduce the annotation burden. 
The co-training framework~\cite{chen2021semi, wang2022cnn, wang2023dual} has attracted widespread attention in SSL due to its ease of implementation and high compatibility. 
Recent trends~\cite{wang2013co, ma2024semi} demonstrate that leveraging the diversity in co-training appropriately helps prevent inferior network performance, capitalizing on the strengths of each network to compensate for the weaknesses of the other. For example, \cite{ma2024semi} has shown excellent performance through co-training with CNN- and SSM-based U-Net networks, exploiting the diversity in network architectures. 

We surprisingly find that in SSM-based networks, the scanning mechanism of the SS2D module can indeed produce diversity by scanning the inputs from different directions as Figure~\ref{fig: intro}, which has not been explored in previous works. This observation inspires us to explore the potential of existing and yet undiscovered diversities regarding SSM-based networks.  
Therefore, we wonder, \textit{is it possible to effectively utilize the diversity to improve the performance of SSM-based semi-supervised medical image segmentation?}

To achieve this goal, we first systematically analyze it from three distinct perspectives: data, network, and feature.

\begin{figure}[t]
   \centering
   \includegraphics[width=0.9\linewidth]{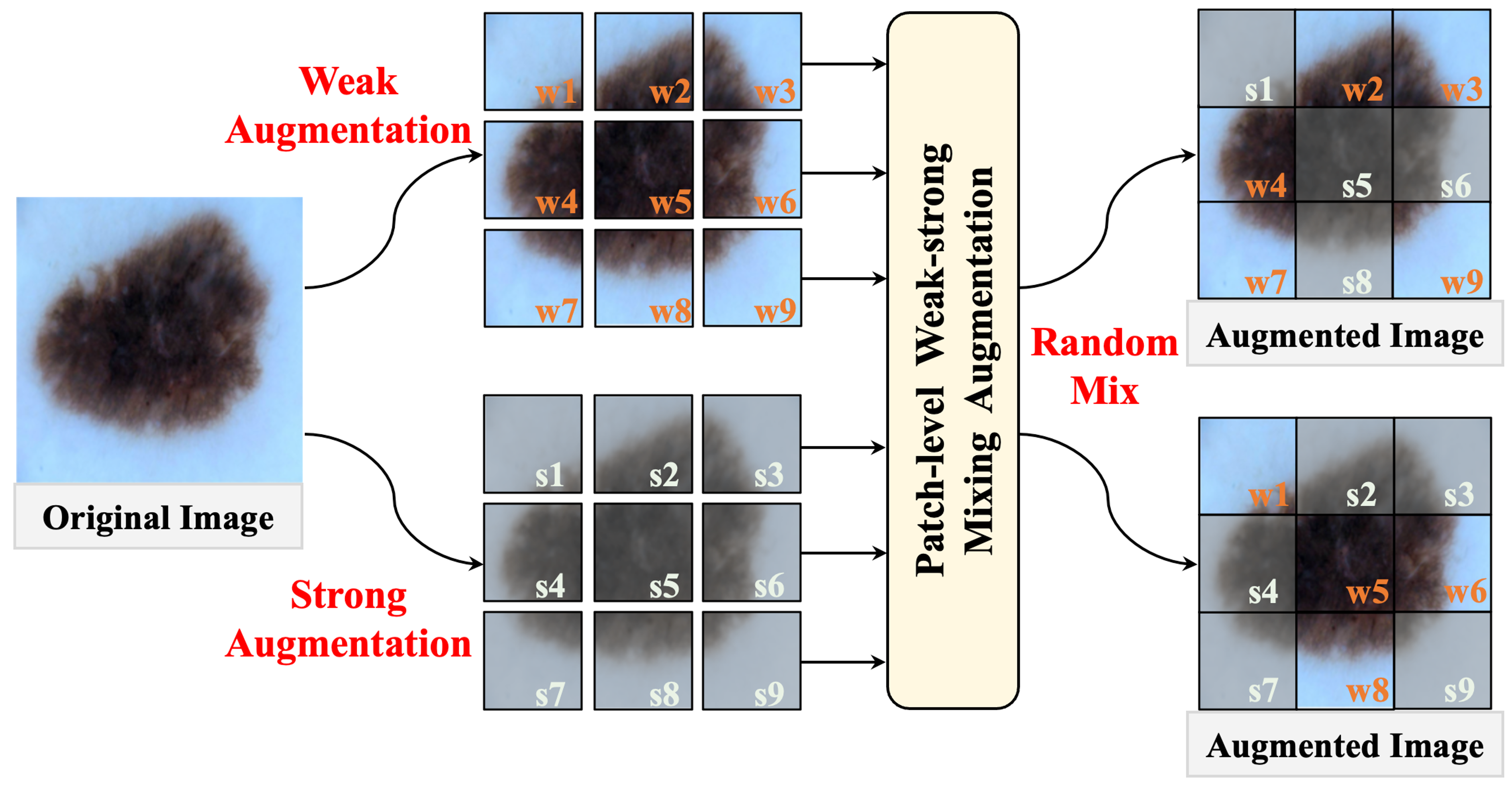}
   \caption{Our patch-level weak-strong mixing augmentation. The strong and weak augmentations of each patch are randomly mixed. It enhances the diversity of local patches while preserving the overall structural integrity. Refer to Section~\ref{sec:diverse-aug} for more details. }
   \label{fig: diverse_aug}
\end{figure}

\textbf{Data perspective: Diversity in patch-level augmentation. }
Mamba in visual tasks divides an image into patches and processes image patches as a sequence~\cite{liu2024vmamba}. This processing way inspires us to introduce diversity among these patches, potentially encouraging diversity in representation over the same sequence of patches, as shown in Figure~\ref{fig: diverse_aug}. 
By varying weak-strong mixing augmentation applied to different local patches, the network is exposed to a wider range of transformations and variations within the input data, enhancing the model's robustness to local feature changes. 
Additionally, patch-level augmentation maintains the overall structural integrity and preserves contextual information of patch sequences. 

\textbf{Network perspective: Diversity in scanning directions. }
We observe that different scanning directions could produce prediction divergence. The SS2D module enables the network to accumulate the history along the scanning route~\cite{liu2024vmamba} and has a strong perception of information in certain directions. The spatial features of targets in medical images, especially lesions, may be distributed in specific directions. 
We explore the effects of different scanning axes on medical images with a network that scans in horizontal and vertical directions and a network that scans in diagonal and anti-diagonal directions. 
The former outperforms the latter by 5\% on vertical targets (as shown in the right of Figure~\ref{fig: intro}), while the diagonal and anti-diagonal scanning exhibits a notably superior performance of 19\% on tilted lesions (as in the left of Figure~\ref{fig: intro}). The results show that the scanning direction greatly affects the segmentation performance, which has not been explored by previous works. This drives us to enhance diversity in semi-supervised frameworks using Mamba's specific scanning modeling. 

\textbf{Feature perspective: Diversity in feature representations. }
We notice that the feature representations of different networks should be distinct from each other to boost the performance of their predictions~\cite{wang2007analyzing, xia20203d}. Consequently, our objective is to enforce constraints on the learning process by encouraging diverse feature learning while preventing the extraction of similar or redundant features. To promote differentiation in feature representations across different networks, we employ uncertainty-weighted feature weighting to encourage the extraction of representative features. This strategy helps the model learn diverse feature representations, thereby enhancing the model's performance and generalization capabilities. 

By exploring and utilizing the diversity in data, network, and feature, we propose a unified Diversity-enhanced Collaborative Mamba framework (DCMamba) for semi-supervised medical image segmentation.
Our framework is intended to fully capitalize on diversity to harness the potential of cross-supervision between two branches.
We first develop patch-level weak-strong mixing augmentation to enhance the diversity of data.
Subsequently, guided by the scanning modeling characteristics of VMamba~\cite{liu2024vmamba}, we introduce a diverse-scan collaborative module. This module employs two SSM-based U-Net networks with distinct scanning directions to model the same sample.
In addition, we propose a diverse-feature fusion mechanism, which adopts uncertainty-weighted contrastive learning to increase the diversity of feature representation of two networks.
Extending beyond 2D tasks, we further extend our method to 3D segmentation tasks by adapting the diversity-enhanced strategies and collaborative mechanisms to 3D backbone networks.
Our main contributions could be summarized as follows: 
\begin{itemize}
    \item We propose a novel diversity-enhanced semi-supervised medical image segmentation framework DCMamba, that incorporates diversity considerations from three perspectives - data, network, and feature. 
    \item Our patch-level weak-strong mixing augmentation aligns well with patch-based modeling of visual SSMs, and encourages data diversity across different patches. 
    \item Our diverse-scan collaborative module fully exploits the sequential processing capabilities of Mamba, cross-supervising networks with different scanning directions. 
    \item Our simple yet effective diverse-feature fusion mechanism efficiently enhances feature representation diversity through uncertainty-weighted contrastive learning.
\end{itemize}
We conduct comprehensive experiments to demonstrate the potential of our DCmamba, which achieves remarkable improvements for semi-supervised segmentation on both single-target and multi-target medical image datasets. Our method outperforms the latest SSM-based method~\cite{ma2024semi} by 6.69\% on the Synapse dataset with 20\% labeled data. Furthermore, we extend DCMamba to 3D segmentation tasks, leveraging spatial context while preserving the advantages of selective scanning.

\section{Related Work}
\subsection{State Space Models for Medical Images}
In recent years, most medical image segmentation methods are primarily based on the architecture of CNNs~\cite{ronneberger2015u} and Transformers~\cite{cao2022swin}. Some approaches~\cite{luo2022semi, wang2022cnn, huang2024combinatorial} exploit the complementary strengths of CNNs and Transformers.
3D-Vit~\cite{wang2023dual} designs a Dual-Transformer architecture, surpassing the combination of CNN and Transformer under certain conditions. FRCNet~\cite{he2024frcnet} integrates frequency domain consistency and region similarity consistency in Dual-Transformer architecture. However, CNN-based networks exhibit certain limitations in capturing long-range dependencies due to the local receptive fields. Transformer-based networks could model global information, but their self-attention mechanism causes computational complexity to grow quadratically with the increase in context length. Both of them face challenges in effectively handling long-range dependencies in medical images.

Recently, SSMs~\cite{mehta2023long, wang2023selective}, exemplified by Mamba~\cite{gu2023mamba}, have been widely studied in the field of medical image segmentation~\cite{ma2024u, ruan2024vm, wang2024mamba, yang2024cardiovascular, liu2024swin}. They demonstrate the potential for modeling long-range interactions and maintain linear computational complexity. 
U-Mamba~\cite{ma2024u} designs a hybrid CNN-SSM block that combines SSM with the local feature extraction capability of convolutional layers. The convolutional features are flattened into a 1D sequence and then processed by SSM block. To alleviate the problem of restricted receptive fields of unidirectional scanning, VMamba~\cite{liu2024vmamba} introduces the SS2D module, which adopts a four-way scanning strategy to enable each element in a feature map to integrate information from all other positions in different directions. Inspired by this, VM-UNet~\cite{ruan2024vm} and Mamba-Unet~\cite{wang2024mamba} integrate the SS2D module of VMamba into U-shaped architecture, to learn spatial information across different scales.
Meanwhile, some other studies~\cite{zhao2024rs, fan2024slicemamba} focus on improving the scanning mechanism. 

There are limited studies on SSM-based networks for semi-supervised segmentation. Semi-Mamba-UNet~\cite{ma2024semi} utilizes Mamba-Unet~\cite{wang2024mamba} and U-Net~\cite{ronneberger2015u} for co-training. 
However, the potential of SSM-based networks and the in-depth analysis of their diversity has not been fully explored. 

\subsection{Semi-supervised Medical Image Segmentation}
To alleviate the scarcity of labeled data, semi-supervised medical image segmentation enhances model representation learning by designing supervision signals for a substantial amount of unlabeled data. 
Common SSL methods can be broadly divided into two main categories: entropy minimization based methods~\cite{lee2013pseudo, yang2022st++} and consistency regularization based methods~\cite{ouali2020semi, luo2021semi, luo2022semi, bai2023bidirectional, miao2023caussl, yang2023revisiting, liu2023semi, huang2024combinatorial, he2024frcnet, huang2024exploring}. Entropy minimization based methods, popularized by self-training~\cite{lee2013pseudo}, focus on selecting pseudo-labeled samples to guide the training process, while consistency regularization based methods aim to produce consistent predictions for different perturbed views of the same instance.
As an extension of self-training, co-training~\cite{chen2021semi, wang2022cnn, wang2023dual} introduces the notion of different individual networks benefiting from each other. 

To help the model better adapt to unlabeled data to improve segmentation performance, some research focuses on generating prediction disagreement under perturbations.
The strong-weak augmentation consistency concept from FixMatch~\cite{sohn2020fixmatch} has been applied to medical images. BCP~\cite{bai2023bidirectional} copypastes random crops from labeled images with random crops from unlabeled images, to align the empirical distribution of labeled and unlabeled features. UniMatch~\cite{yang2023revisiting} involves both weak and strong augmentations and network perturbations.
Based on the network perturbation of the teacher-student structure~\cite{tarvainen2017mean}, DTC~\cite{luo2021semi} introduces an additional task-level constraint, and ACMT~\cite{xu2023ambiguity} introduces an ambiguous target selection strategy. 
CCT~\cite{ouali2020semi} perturbs the features of the encoder and performs consistency learning of multiple decoders. 
However, these techniques are all based on CNN or Transformer networks. We explore these important factors, \eg, weak-strong augmentation and consistency learning, within semi-supervised segmentation in advanced SSM-based networks. 

\begin{figure}[t]
   \centering
   \includegraphics[width=1.0\linewidth]{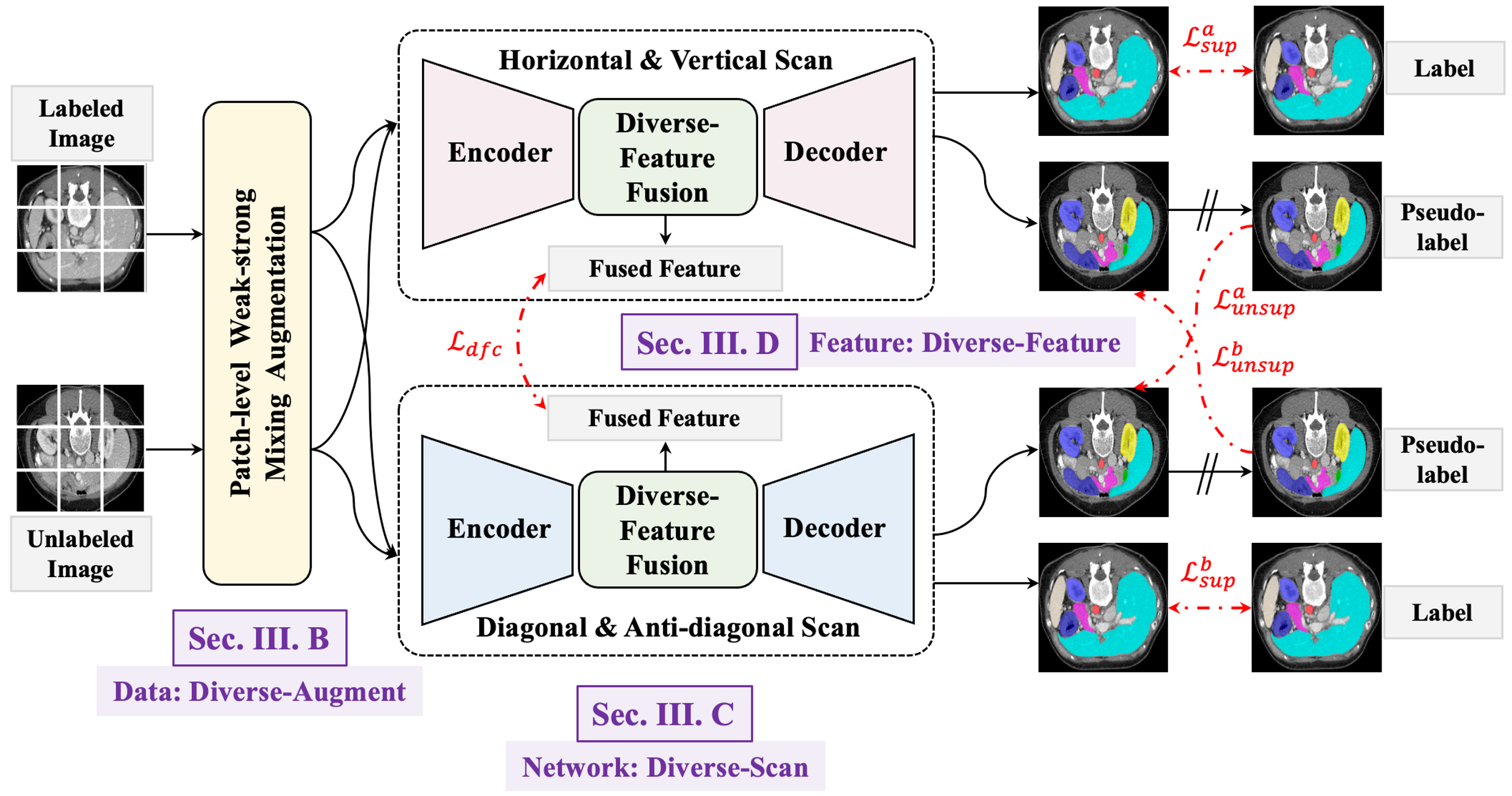}
   \caption{Overview of our proposed framework. Our framework enhances diversity from the perspectives of data, network, and feature to harness the potential of cross-supervision between two branches. We introduce a strategy for patch-level weak-strong mixing augmentation, followed by feeding two augmented samples into two SSM-based networks with different scanning directions. The diverse-feature fusion mechanism employs uncertainty-weighted contrastive learning. }
   \label{fig: framework}
\end{figure}

\section{Method}
\begin{figure}[t]
   \centering
   \includegraphics[width=0.75\linewidth]{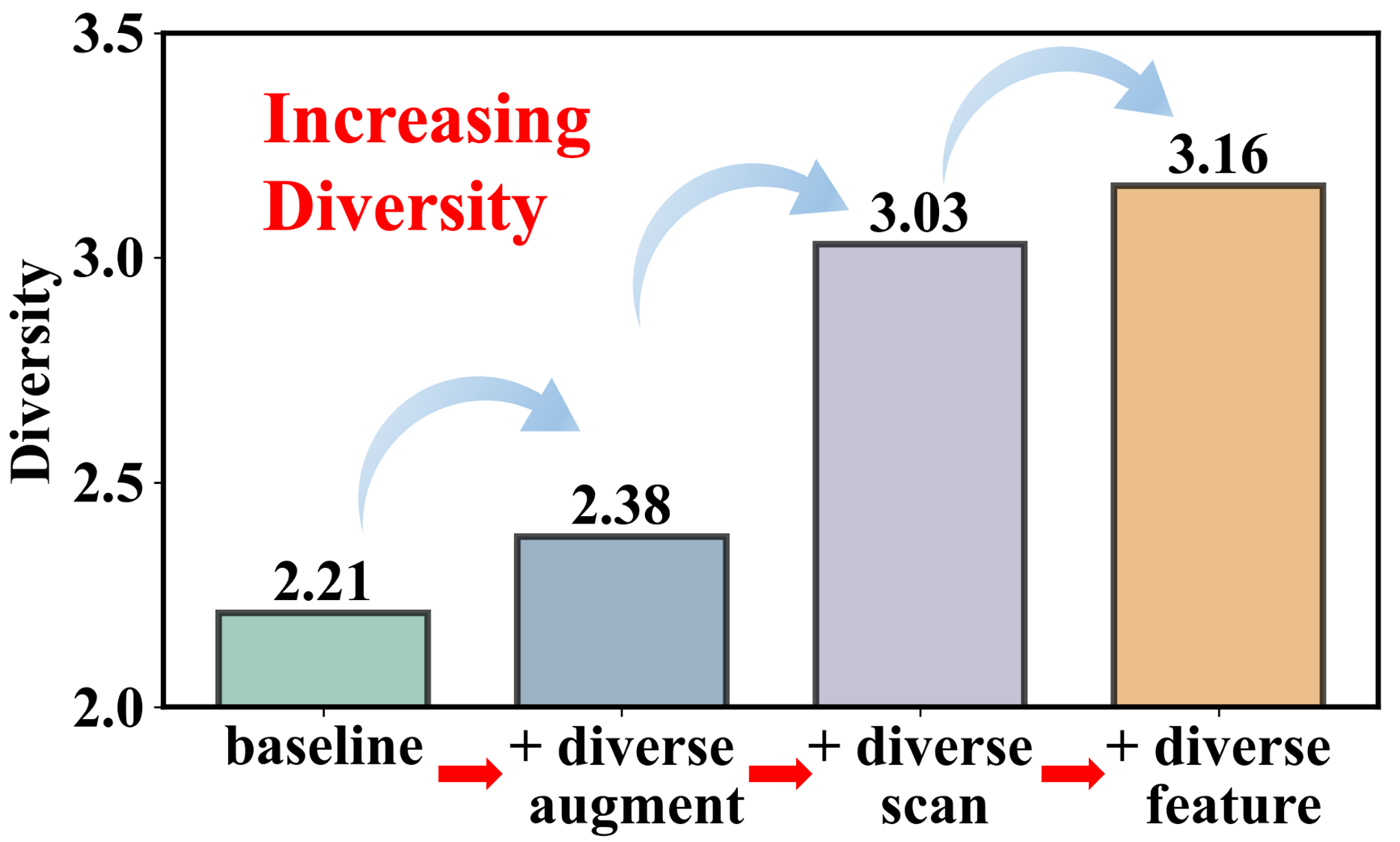}
   \caption{Our method effectively increases diversity. We calculate the cosine distance between the feature representations of two networks on the testing set and use it as the diversity measure. It enhances diversity through patch-level weak-strong mixing augmentation (diverse-augment), diverse-scan collaborative Mamba (diverse-scan), and a diverse-feature fusion mechanism (diverse-feature). }
   \label{fig: diversity_increase}
\end{figure}

In this work, we propose a novel semi-supervised medical image segmentation framework, DCMamba, designed to explore diversity in data, networks, and feature representations. The overall framework is illustrated in Figure~\ref{fig: framework}.
In our framework, we develop a patch-level weak-strong mixing augmentation strategy and then input two augmented samples into different networks respectively (in Section~\ref{sec:diverse-aug}). The diverse-scan collaborative module is composed of two SSM-based networks with different scanning directions (in Section~\ref{sec:diverse-scan}). We present a diverse-feature fusion mechanism, which employs uncertainty-weighted contrastive learning to capture extensive feature patterns across the two networks (in Section~\ref{sec:diverse-feat}). 
Additionally, we extend our framework to 3D medical image segmentation tasks by adapting the diverse-augment, diverse-scan, and diverse-feature mechanisms to 3D networks (in Section~\ref{sec:method3d}).
As shown in Figure~\ref{fig: diversity_increase}, our framework enhances diversity from the perspectives of data, network, and feature. 

\subsection{Preliminaries}
Inspired by linear time-invariant systems, SSM~\cite{mehta2023long, wang2023selective} maps the input sequence $x_t \in \mathbb{R}$ to response sequence $y_t \in \mathbb{R}$, through a latent state representation $h \in \mathbb{R}^{N}$. It can be represented as the subsequent linear ordinary differential equations: 
\begin{equation}
h'_t = Ah_t + Bx_t,\quad y_t = Ch_t,
\end{equation}
where $A \in \mathbb{R}^{N \times N}$ as the evolution parameter and $B \in \mathbb{R}^{N \times 1}$, $C \in \mathbb{R}^{1 \times N}$ as the projection parameters.
The discrete SSM discretizes this continuous system to make it more suitable for deep learning scenarios, introducing a timescale parameter $\Delta$ and transforming $A$ and $B$ into discrete parameters $\bar{A}$ and $\bar{B}$ using a fixed discretization rule, which can be defined as follows:
\begin{equation}
\bar{A} = e^{\Delta A}, \bar{B} = (\Delta A)^{-1}(e^{\Delta A} - I) \cdot \Delta B,
\end{equation}

Mamba~\cite{gu2023mamba} further improves the SSM by introducing a selective scan mechanism, enabling efficient information filtering by dynamic parameter adjustments based on the context of the input sequence.
To process the vision tasks, VMamba~\cite{liu2024vmamba} introduces the SS2D module to arrange image patches in four different routes to generate separate sequences, and integrate information of the sequences.

\subsection{Patch-level Weak-strong Mixing Augmentation}
\label{sec:diverse-aug}
In the semi-supervised study, the training dataset consists of the labeled data $\mathcal{D}_L = \left\{\left(x_i, y_i \right) \right\}_{i = 1, \ldots, l}$ and the unlabeled data $\mathcal{D}_U = \left\{X_i\right\}_{i = l+1, \ldots, n}$, where $x_i \in \mathbb{R}^{H \times W}$ is the input image and $y_i \in \left\{0, 1, ..., c\right\}^{H \times W}$ represents ground truth with $c$ number of classes. 
Our method employs a diverse augmentation strategy by separately feeding patch-level weak-strong mixing augmented versions of both labeled and unlabeled data into distinct networks. 
Specifically, as shown in Figure~\ref{fig: diverse_aug}, we first split image $x_i$ into $d \times d$ patches (taking $d=3$ as an example), $x_i = [p_{i,1}, p_{i,2}, ..., p_{i,d^2}]$. The augmented samples are denoted as $x^{\prime}_i$ and $x^{\prime\prime}_i$. Let $A_w$ represent the weak augmentation and $A_s$ represent the strong augmentation functions. For every patch $p_{i,j}$, we perform weak and strong augmentation, $A_w(p_{i,j})$ and $A_s(p_{i,j})$, respectively. Subsequently, $A_w(p_{i,j})$ and $A_s(p_{i,j})$ are randomly placed into the augmented images $x^{\prime}_i$ and $x^{\prime\prime}_i$, ensuring a mix of weak or strong transformations of different patches in both augmented images. This strategy increases the diversity of the two views of co-training from the data level. 

Our patch-level weak-strong mixing augmentation method randomly selects weak or strong augmentations and does not change the shape and structure of the targets. 
Some recent approaches based on CutMix~\cite{yun2019cutmix, bai2023bidirectional} and MixUp~\cite{zhang2018mixup} techniques combine multiple samples to generate new samples. However, these methods may mix unimportant image regions, adversely affecting target recognition. Our approach enhances image diversity at the patch level without distorting the overall structure, while preserving contextual information between sequences of patches. 

\subsection{Diverse-Scan Collaborative Mamba}
\label{sec:diverse-scan}
Our segmentation network architecture follows Mamba-Unet~\cite{wang2024mamba}, which integrates the visual state space (VSS) blocks of VMamba~\cite{liu2024vmamba} on a U-shaped encoder-decoder configuration. Specifically, in the VSS block, the input features are processed through SS2D module, which unfolds the input image along four different directions into sequences. The features extracted from these sequences ensure that information from various directions is thoroughly scanned, and merged sequences from the four directions. SS2D helps capture complex spatial relationships and provides a comprehensive context understanding.

\begin{figure*}[t]
   \centering
   \includegraphics[width=0.85\linewidth]{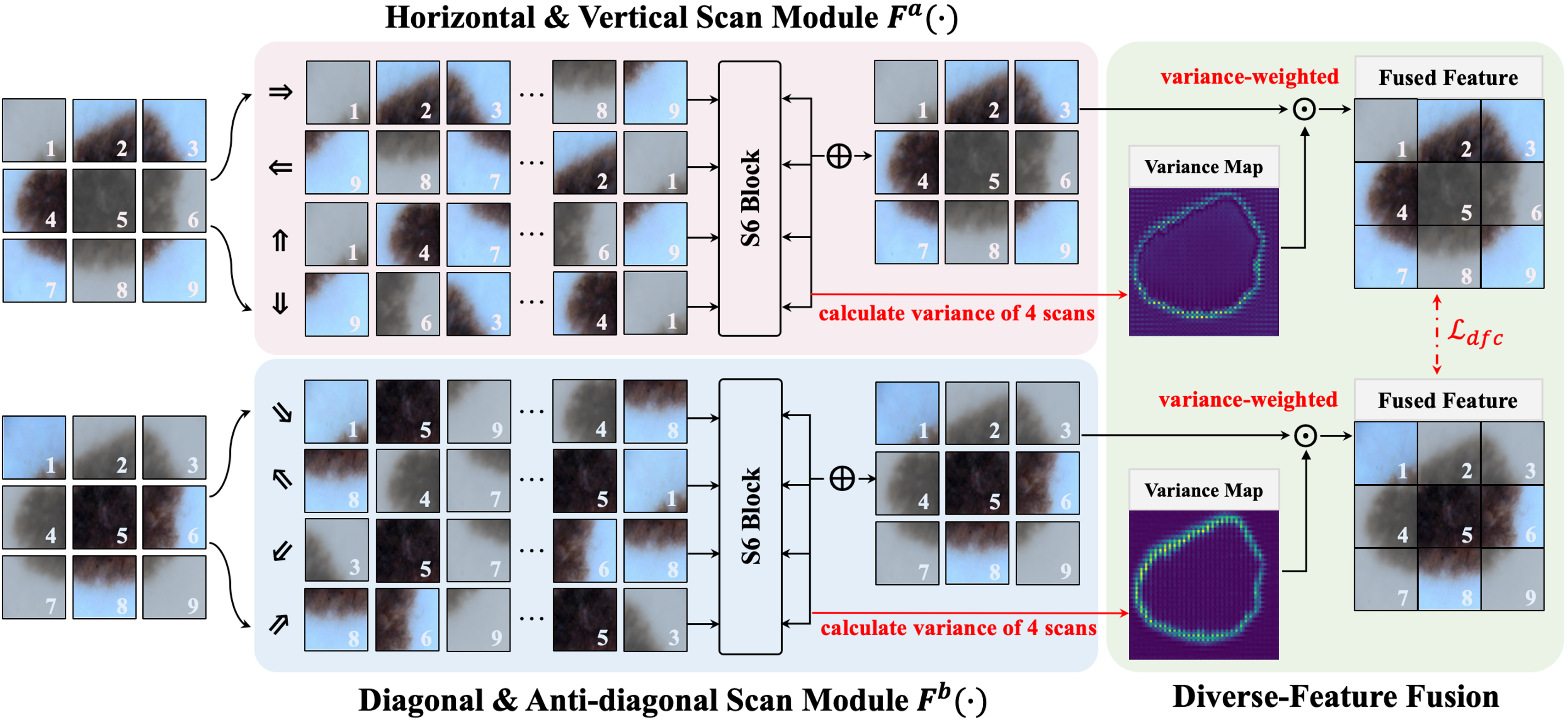}
   \caption{Diverse-scan collaborative module and diverse-feature fusion. The horizontal \& vertical scan module performs selective scanning on the features in the forward and backward directions along the horizontal and vertical directions, and the diagonal \& anti-diagonal scan module selectively scans along the diagonal and anti-diagonal directions. In each module, four scanning routes are fused using pixel-level variance weighting.}
   \label{fig: diverse_scan_feat}
\end{figure*}

The original SS2D module is designed to model along horizontal and vertical directions, while medical images are non-sequential and their spatial features are often present in arbitrary directions. By utilizing different scanning directions in diverse manners to learn the feature representation, it is possible to capture distinct features. To enhance network diversity, we propose the diverse-scan collaborative Mamba. In the two networks, we introduce two SS2D modules with different scan directions respectively, as shown in Figure~\ref{fig: diverse_scan_feat}. They both scan the image from top-left to bottom-right, bottom-right to top-left, top-right to bottom-left, and bottom-left to top-right. 
The difference is that the \textit{horizontal \& vertical scan module} performs selective scanning on the features in the forward and backward directions along the horizontal and vertical directions, while the \textit{diagonal \& anti-diagonal scan module} selectively scans along the diagonal and anti-diagonal directions. 
Each of them processes patch sequences independently and performs modeling in a specific direction. The two Mamba networks with diverse scanning are denoted by $F^a(\cdot)$ and $F^b(\cdot)$ respectively. Through collaborative training between two networks, the framework could merge knowledge from multiple directions to adapt the diverse spatial features in medical images, further enriching feature learning. 

\begin{figure}[t]
   \centering
   \includegraphics[width=0.98\linewidth]{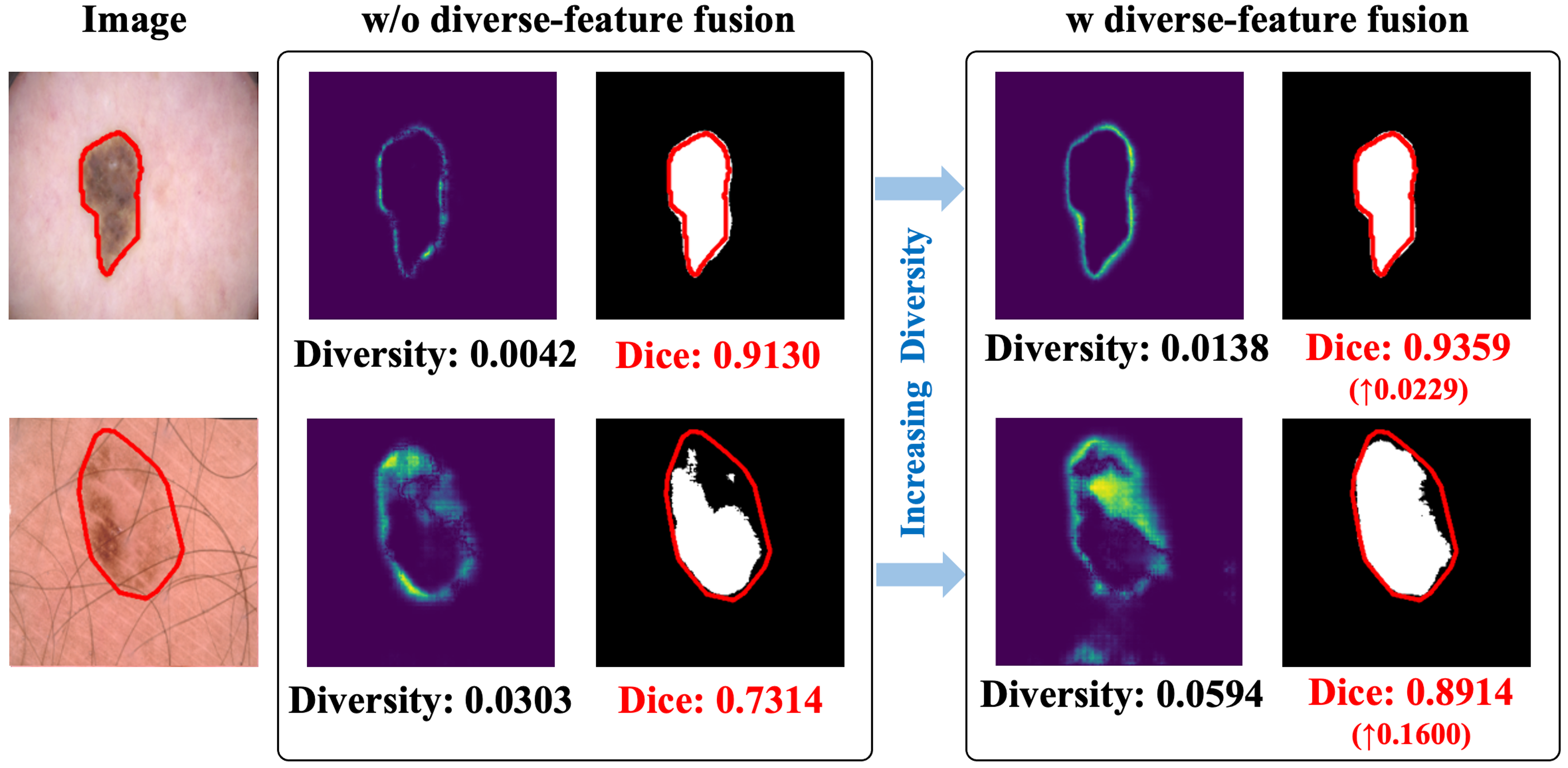}
   \caption{Diverse-feature fusion mechanism can effectively improve the diversity between the two predictions and further improve the segmentation results of the regions with high diversity. We use the absolute value of the difference between the features of the two networks as the measure of diversity. The red curves denote the ground truth. }
   \label{fig: diverse_feat}
\end{figure}

\subsection{Diverse-Feature Fusion}
\label{sec:diverse-feat}
Our two networks with independent scanning directions generate pseudo-labels for mutual supervision. To leverage the collaborative advantages of cross supervision, we propose a diverse-feature contrastive learning module with uncertainty-weighted feature fusion, which effectively improves the diversity as shown in Figure~\ref{fig: diverse_feat}. After adding diverse-feature fusion, there is a noticeable enhancement in the diversity of the two networks, leading to significant improvements in segmentation performance in areas with high diversity, such as edge areas.

Specifically, by assigning pixel-level weights to the feature representations extracted by each network, we enable them to concentrate on learning uncertain regions and capture diverse features. 
As we mentioned before, there are four scanning routes starting from different positions in each network, and producing four embedding sequences. They are added together in each SS2D for fusion. 
In each network, we first extract the feature representations $\left\{ z^1_i, z^2_i, z^3_i, z^4_i \right\}$ of the four routes of the two network encoders respectively, and fuse them into new features $h_i$. Specifically, we utilize the pixel-level variance between predictions from the four routes as a measure of uncertainty, mapping it to a range between 0 and 1 through a logistic function. The pixel-level weights of the network's fused feature map are defined as: 
\begin{equation}
w_i = \frac{1}{1 + \exp(-\frac{1}{\emph{K}} \sum_{k=1}^{\emph{K}} (z^k_i - \bar{z_i})^2)},
\end{equation}
where $\emph{K}$ represents the number of scan routes, and $\bar{z_i}$ is the average feature values of all routes. According to this definition, for a given pixel in the fused feature map, higher uncertainty automatically results in a higher weight. After uncertainty-weighted feature enhancement, the fused feature could be expressed as: 
\begin{equation}
h_i = \left( \sum_{k=1}^{\emph{K}} z^k_i \right) \odot w_i,
\end{equation}
where $\odot$ denotes the element-wise product. 

In our proposed module, a pair of projectors $(P^a, P^b)$ is appended to the encoders of two separate networks to extract feature representations, which are subsequently used to measure the image similarity within an embedding space. We adopt feature contrastive learning to construct an optimized embedding space to minimize the distance between feature representations of the same image with different augmentations while maintaining dissimilarity with the feature representations of different images. The contrastive learning loss between the fused features from the two networks is defined as: 
\begin{equation}
\mathcal{L}_{dfc} = - \sum_{i=1}^{n} \log \frac{e^{\mathrm{sim}\bigl(P^a(h^a_i), P^b(h^b_i)\bigr) / \tau}}{\sum_{x_j \in N^-} e^{\mathrm{sim}\bigl(P^a(h^a_i), P^b(h^b_j)\bigr) / \tau}},
\end{equation}
where $h^a_i$ and $h^b_j$ are the fused features produced by networks $F^a(\cdot)$ and $F^b(\cdot)$, respectively. The fused features of the same image from different networks are considered positive pairs, while features of different images are negative pairs. The set $N^-$ consists of all images other than $x_i$, including their all transformations. The similarity in the representation space is measured by the dot product, \ie, $\mathrm{sim}(a, b) = a \cdot b$ and, $\tau$ is a temperature scaling factor. 

\begin{figure*}[t]
    \centering
    \includegraphics[width = 0.85\textwidth]{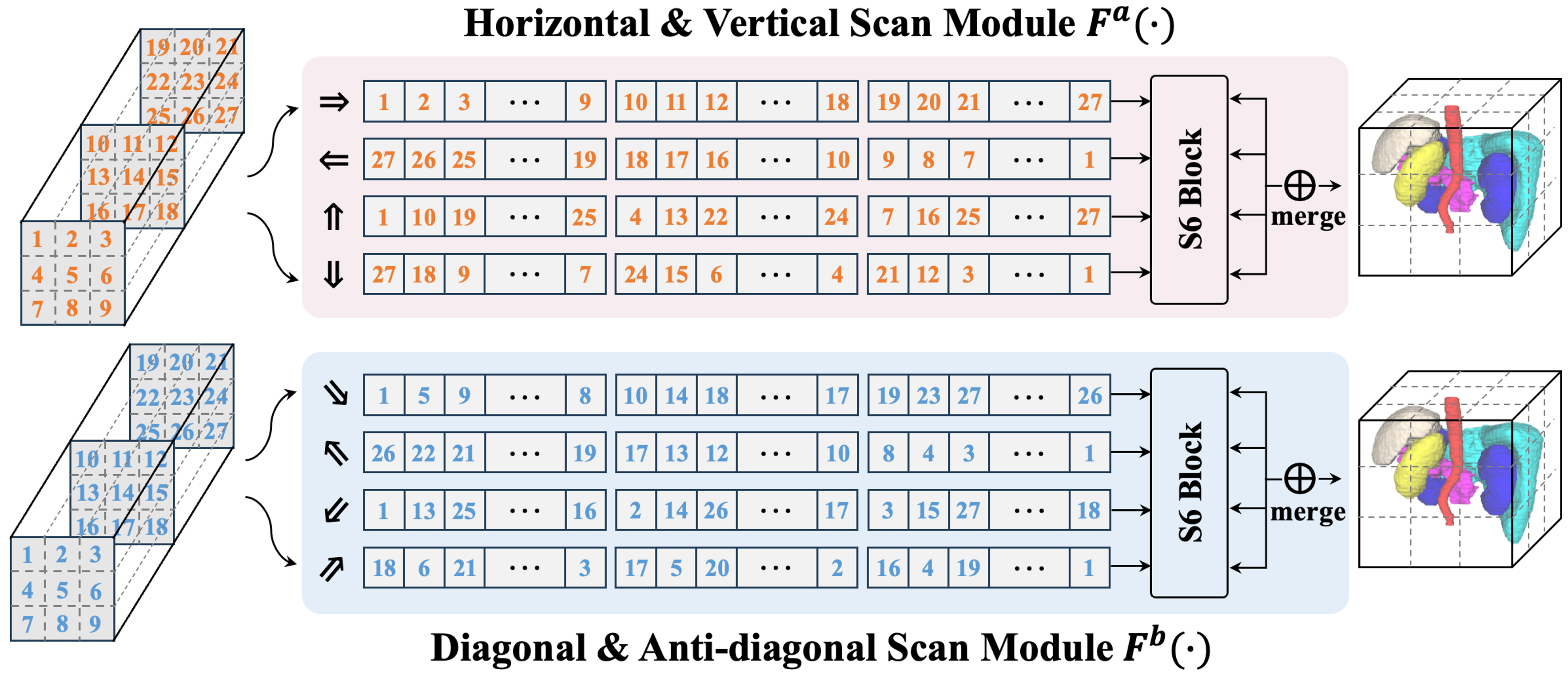}
    \caption{Diverse-scan collaborative module for 3D images.}
    \label{fig: ss3d}
\end{figure*}

\subsection{Training Objective}
Formally, the supervised segmentation loss function with training set is defined as: 
\begin{equation}
\begin{aligned}
\mathcal{L}_{sup} &= \mathcal{L}^a_{sup} + \mathcal{L}^b_{sup} \\
&= \frac{1}{2} \bigl(\mathcal{L}_{dice} (F^a(x^{\prime}_i), y_i) + \mathcal{L}_{ce}(F^a(x^{\prime}_i), y_i)\bigr) \\
&+ \frac{1}{2} \bigl(\mathcal{L}_{dice} (F^b(x^{\prime\prime}_i), y_i) + \mathcal{L}_{ce}(F^b(x^{\prime\prime}_i), y_i)\bigr),
\end{aligned}
\end{equation}
where $\mathcal{L}_{dice}$ and $\mathcal{L}_{ce}$ are dice loss and cross-entropy loss, respectively. 
Following the common practice in semi-supervised learning~\cite{yu2019uncertainty,bai2023bidirectional,ma2024semi}, we use a factor of $\frac{1}{2}$ to balance the cross-entropy loss and dice loss.
$x^{\prime}_i$ and $x^{\prime\prime}_i$ are the weakly augmented and strongly augmented versions of $x_i$, and $(x_i, y_i) \in D_L$. And the cross-supervision loss is illustrated as: 
\begin{equation}
\begin{aligned}
\mathcal{L}_{unsup} &= \mathcal{L}^a_{unsup} + \mathcal{L}^b_{unsup} \\
&= \frac{1}{2} \bigl(\mathcal{L}_{dice} (F^a(x^{\prime}_i), P^b_i) + \mathcal{L}_{ce}(F^a(x^{\prime}_i), P^b_i)\bigr) \\
&+ \frac{1}{2} \bigl(\mathcal{L}_{dice} (F^b(x^{\prime\prime}_i), P^a_i) + \mathcal{L}_{ce}(F^b(x^{\prime\prime}_i), P^a_i)\bigr),
\end{aligned}
\end{equation}
where the prediction of two networks can be considered as $P^a_i = \arg\max(F^a(x^{\prime}_i))$ and $P^b_i = \arg\max(F^b(x^{\prime\prime}_i))$.
The overall objective is minimising the total loss $\mathcal{L}$, which is expressed as: 
\begin{equation}
\mathcal{L} = \mathcal{L}_{sup} + \lambda \mathcal{L}_{unsup} + \mathcal{L}_{dfc},
\end{equation}
where $\lambda$ is a time-dependent Gaussian warming up function, which is defined as $\lambda\left(t\right) = 0.1 \cdot e^{-5{\left(1 - \frac{t}{t_{max}} \right)}^2}$, where $t$ and $t_{max}$ denote the current and maximum training iterations, respectively. The use of a dynamic weight factor during training~\cite{laine2016temporal,tarvainen2017mean,yu2019uncertainty} is to balance the contribution of labeled and unlabeled data over time.

\subsection{The 3D Extension of DCMamba}
\label{sec:method3d}
To further enhance our framework, we extend it to 3D tasks. The core principles of our method, including patch-level weak-strong mixing augmentation (diverse-augment), the diverse-scan collaborative module (diverse-scan), and the diverse-feature fusion mechanism (diverse-feature), are highly adaptable to 3D networks. These mechanisms could be naturally extended to the 3D domain, where they leverage the spatial information of the context while maintaining the advantages of selective scanning.

Specifically, we integrate a 3D visual state space (3D VSS) block on an encoder-decoder structure with skip connections. 3D VSS has a similar architecture to the 2D VSS block of VMamba~\cite{liu2024vmamba}, and extending the depth-wise convolution to 3D convolution enables the network to utilize three-dimensional spatial information. In the 3D VSS block, the input features are processed through the 3D selective scan module, which also similar to SS2D, unfolds the input image along four different directions into sequences. The features extracted from these sequences ensure that information from various directions is thoroughly scanned, and merged sequences from the four directions. In our DCMamba framework, we design two scan modules with different scan directions respectively for two networks, as shown in Figure~\ref{fig: ss3d}. $F^a(\cdot)$ is a scan module that scans in horizontal and vertical directions and $F^b(\cdot)$ is in diagonal and anti-diagonal directions.

We extend our framework to 3D tasks by applying the weak-strong mixing augmentation method on 3D patches and incorporating uncertainty-weighted feature fusion for 3D diverse-scan collaborative networks.

\section{Experiments}

\begin{table*}[t]
\caption{Results of Synapse dataset with 10\% and 20\% annotations. ``L/U" indicates the labeled and unlabeled samples.}
\label{tab: synapse}
\centering
\scalebox{0.85}{
\begin{tabular}{c|c|c|cc|cccccccc}
\hline
\noalign{\smallskip}
\multirow{2.5}{*}{L / U} & \multirow{2.5}{*}{Method} & \multirow{2.5}{*}{Venue} & \multirow{2.5}{*}{\begin{tabular}[c]{@{}c@{}}Average \\ Dice (\%) $\uparrow$\end{tabular}} & \multirow{2.5}{*}{\begin{tabular}[c]{@{}c@{}}Average \\ ASD $\downarrow$\end{tabular}} & \multicolumn{8}{c}{Average Dice of Each Class (\%) $\uparrow$} \\ 
\noalign{\smallskip}
\cline{6-13}
\noalign{\smallskip}
& & & & & Arota & Gallbladder & R.Kedney & L.Kedney & Liver & Pancreas & Spleen & Stomach  \\ 
\noalign{\smallskip}
\hline
\noalign{\smallskip}
\multirow{10}{*}{\begin{tabular}[c]{@{}c@{}}4 / 14 \\ (20 / 80\%)\end{tabular}}
& Mean Teacher\scalebox{0.7}{~\cite{tarvainen2017mean}} & NIPS'17 & 45.29$\pm$2.87 & 34.61$\pm$5.81 & 67.71 & 29.11 & 44.08 & 40.02 & 72.43 & 14.85 & 55.56 & 38.56 \\
& Fixmatch\scalebox{0.7}{~\cite{sohn2020fixmatch}} & NIPS'20 & 60.35$\pm$4.07 & 17.61$\pm$5.73 & \textbf{79.72} & \textbf{45.03} & 57.87 & 64.47 & 80.45 & 30.10 & 75.42 & 49.72 \\
& CPS\scalebox{0.7}{~\cite{chen2021semi}} & CVPR'21 & 46.54$\pm$3.48 & 28.51$\pm$2.06 & 68.42 & 32.34 & 42.46 & 45.00 & 75.41 & 15.59 & 55.80 & 37.30 \\
& SS-Net\scalebox{0.7}{~\cite{wu2022exploring}} & MICCAI'22 & 51.07$\pm$4.12 & 24.68$\pm$4.65 & 72.71 & 35.20 & 51.26 & 50.78 & 76.76 & 13.13 & 66.85 & 41.83 \\
& ACMT\scalebox{0.7}{~\cite{xu2023ambiguity}} & MedIA'23 & 44.93$\pm$4.17 & 30.99$\pm$7.71 & 66.02 & 26.77 & 41.77 & 44.98 & 73.77 & 15.35 & 51.29 & 39.53 \\
& BCP\scalebox{0.7}{~\cite{bai2023bidirectional}} & CVPR'23 & 60.25$\pm$1.89 & 25.50$\pm$10.61 & 77.28 & 25.06 & 72.19 & 62.74 & 79.38 & \textbf{37.36} & 75.25 & 52.73 \\
& IC\scalebox{0.7}{~\cite{huang2024exploring}} & TMI'24 & 58.29$\pm$2.36 & 15.09$\pm$1.12 & 74.50 & 35.82 & 71.11 & 67.48 & 88.02 & 10.67 & 79.19 & 39.54 \\
& CML\scalebox{0.7}{~\cite{wu2024cross}} & ACM MM'24 & 44.24$\pm$5.16 & 46.17$\pm$9.99 & 71.70 & 9.85 & 51.97 & 57.70 & 67.74 & 4.21 & 61.93 & 28.80 \\
& Semi-Mamba-UNet\scalebox{0.7}{~\cite{ma2024semi}} & Arxiv'24 & 60.58$\pm$3.61 & 10.99$\pm$5.90 & 70.54 & 25.74 & 71.64 & 65.67 & 82.87 & 35.79 & 78.20 & 54.21 \\
& \cellcolor{gray!20}\textbf{DCMamba (Ours)} & \cellcolor{gray!20}- & \cellcolor{gray!20}\textbf{67.27$\pm$0.81} & \cellcolor{gray!20}\textbf{9.28$\pm$0.74} & \cellcolor{gray!20}75.17 & \cellcolor{gray!20}36.09 & \cellcolor{gray!20}\textbf{80.80} & \cellcolor{gray!20}\textbf{78.28} & \cellcolor{gray!20}\textbf{91.60} & \cellcolor{gray!20}32.52 & \cellcolor{gray!20}\textbf{81.00} & \cellcolor{gray!20}\textbf{62.70} \\
\noalign{\smallskip}
\hline
\noalign{\smallskip}
\multirow{10}{*}{\begin{tabular}[c]{@{}c@{}}2 / 16 \\ (10 / 90\%)\end{tabular}}
& Mean Teacher\scalebox{0.7}{~\cite{tarvainen2017mean}} & NIPS'17 & 35.90$\pm$3.94 & 34.13$\pm$4.04 & 52.57 & 21.09 & 25.95 & 39.28 & 73.49 & 9.44 & 39.76 & 25.65 \\
& Fixmatch\scalebox{0.7}{~\cite{sohn2020fixmatch}} & NIPS'20 & 51.32$\pm$5.13 & 20.57$\pm$4.85 & 75.16 & 21.98 & 45.77 & 53.20 & 86.34 & 18.91 & 67.33 & 41.87 \\
& CPS\scalebox{0.7}{~\cite{chen2021semi}} & CVPR'21 & 35.96$\pm$2.74 & 33.02$\pm$5.48 & 51.62 & 20.72 & 26.73 & 40.53 & 77.17 & 9.37 & 39.47 & 22.09 \\
& SS-Net\scalebox{0.7}{~\cite{wu2022exploring}} & MICCAI'22 & 40.94$\pm$5.68 & 28.18$\pm$10.72 & 66.54 & 16.06 & 34.54 & 44.11 & 80.67 & 9.96 & 47.91 & 27.74 \\
& ACMT\scalebox{0.7}{~\cite{xu2023ambiguity}} & MedIA'23 & 35.38$\pm$2.92 & 35.02$\pm$13.56 & 53.80 & 19.88 & 26.77 & 38.44 & 72.76 & 10.29 & 38.25 & 22.82 \\
& BCP\scalebox{0.7}{~\cite{bai2023bidirectional}} & CVPR'23 & 54.69$\pm$9.07 & 38.94$\pm$22.96 & 68.25 & \textbf{28.72} & 65.25 & 62.29 & 76.49 & \textbf{23.76} & 68.90 & 43.84 \\
& IC\scalebox{0.7}{~\cite{huang2024exploring}} & TMI'24 & 49.46$\pm$3.15 & 14.32$\pm$5.15 & 71.50 & 22.94 & 60.86 & 63.57 & 85.12 & 6.68 & 65.64 & 19.33 \\
& CML\scalebox{0.7}{~\cite{wu2024cross}} & ACM MM'24 & 43.99$\pm$7.41 & 44.84$\pm$22.07 & \textbf{74.39} & 9.84 & 51.48 & 58.32 & 79.30 & 9.84 & 56.71 & 12.04 \\
& Semi-Mamba-UNet\scalebox{0.7}{~\cite{ma2024semi}} & Arxiv'24 & 55.89$\pm$2.43 & 12.88$\pm$0.54 & 63.82 & 19.68 & 69.36 & 68.82 & 87.01 & 23.38 & 74.04 & 40.99 \\
& \cellcolor{gray!20}\textbf{DCMamba (Ours)} & \cellcolor{gray!20}- & \cellcolor{gray!20}\textbf{60.41$\pm$0.84} & \cellcolor{gray!20}\textbf{8.96$\pm$0.70} & \cellcolor{gray!20}69.52 & \cellcolor{gray!20}15.83 & \cellcolor{gray!20}\textbf{79.78} & \cellcolor{gray!20}\textbf{76.54} & \cellcolor{gray!20}\textbf{89.80} & \cellcolor{gray!20}23.03 & \cellcolor{gray!20}\textbf{75.82} & \cellcolor{gray!20}\textbf{52.97} \\
\noalign{\smallskip}
\hline
\end{tabular}}
\end{table*}

\begin{figure*}[t]
   \centering
   \includegraphics[width=1.0\linewidth]{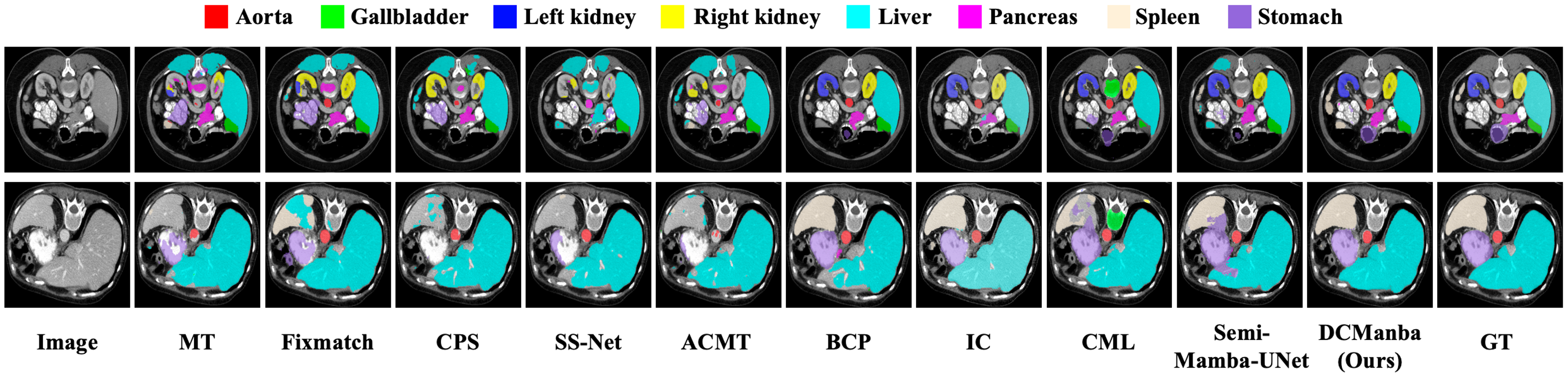}
   \caption{The segmentation results on the Synapse dataset, where GT denotes corresponding ground truth labels. }
   \label{fig: synapse}
\end{figure*}

\subsection{Datasets}
To verify the effectiveness of the proposed method, we conduct comparative experiments on six public medical image datasets, Synapse, ISIC17, ISIC18, Kvasir, AbdomenCT-1K, and FLARE22 datasets. Among them, 2D segmentation experiments are performed on Synapse, ISIC17, ISIC18, and Kvasir, while 3D segmentation experiments are conducted on Synapse, AbdomenCT-1K, and FLARE22.
\subsubsection{Synapse Dataset}
Synapse multi-organ segmentation dataset (Synapse)~\cite{landman2015miccai} is a publicly available multi-organ segmentation dataset. It includes 30 cases with 3,779 axial abdominal clinical CT images, including 8 types of abdominal organs (aorta, gallbladder, left kidney, right kidney, liver, pancreas, spleen, stomach). Following the setting of previous works~\cite{cao2022swin, ruan2024vm}, 18 cases are used for training and 12 cases for testing, and all slices are resized to $224 \times 224$.
We also use the Synapse dataset for 3D tasks and the volume is divided into patches of size 96 $\times$ 96 $\times$ 96.

\subsubsection{ISIC17 and ISIC18 Datasets}
ISIC17~\cite{codella2018skin} and ISIC18~\cite{codella2019skin} are two publicly available skin lesion segmentation datasets published by the International Skin Imaging Collaboration (ISIC). In the experiments, we split the datasets in a 7:3 ratio for use as training and testing sets and resize the images to $256 \times 256$ following the previous work~\cite{ruan2024vm}. For ISIC17, the training set includes 1,500 images, while the test set contains 650 images. For ISIC18, there are 1,886 images for training and 808 images for testing. 

\subsubsection{Kvasir-SEG Dataset}
There are 1,000 images and their corresponding labels in the Kvasir-SEG dataset~\cite{jha2020kvasir} for gastrointestinal polyp segmentation. We randomly select 80\% for training and the remaining for testing. Following the previous work~\cite{wu2024h}, the images and labels are resized to $256 \times 256$.

\subsubsection{AbdomenCT-1K Dataset}
AbdomenCT-1K dataset~\cite{ma2021abdomenct} is a diverse abdominal CT organ segmentation dataset, with more than 1,000 CT scans from 12 medical centers, including multi-phase, multi-vendor, and multi-disease cases. We follow the same pre-processing procedure as nnU-Net~\cite{isensee2021nnu}. We split 900 scans in the training dataset and 100 scans in the test dataset. During training, we randomly cropped patches of size $96 \times 96 \times 96$. In the testing phase, a sliding window approach is applied with the stride of $32 \times 32 \times 32$.

\subsubsection{FLARE22 Dataset}
The FLARE22 dataset~\cite{huang2023revisiting} consists of 13 classes of organs (with one background): the liver (Liv), spleen (Spl), pancreas (Pan), right kidney (R.kid), left kidney (L.kid), stomach (Sto), gallbladder (Gal), esophagus (Eso), aorta (Aor), inferior vena cava (IVC), right adrenal gland (RAG), left adrenal gland (LAG), and duodenum (Duo). We partition the dataset following the previous work~\cite{zhao2024guidednet} with 420 training data and 14 test data. Also, we use the same 42 labeled cases and 378 unlabeled cases. The random crop size is set to $64 \times 128 \times 128$ and the training parameters follow the settings used in previous studies~\cite{zhao2024guidednet}.

\subsection{Implementation Details}
For training cases, weak data augmentations include random flip and random rotation, and strong augmentations consist of pixel-wise transforms (Gaussian blur, brightness, contrast, gamma), following the previous work~\cite{wang2023towards}. For each sample, we set the probability of each transformation being applied to be 0.9. 
For SSM-based networks, our backbone is Mamba-UNet architecture~\cite{wang2024mamba}. We initialize the weights with the VMamba-Tiny~\cite{liu2024vmamba}, which is pre-trained on ImageNet-1k. During the training period, for 2D datasets, the batch size is set to 24 and the labeled batch size is 12. For 3D datasets, the batch size is set to 4 with 2 labeled samples. Following the existing SSL methods~\cite{tarvainen2017mean, cao2022swin}, we employ the SGD optimizer with momentum 0.9 and weight decay 1e-4 to optimize our model for back propagation. We train AbdomenCT-1K dataset for 10k iterations, and all other datasets for 20k iterations with a base learning rate set to 0.01.
All experiments are conducted on Nvidia GeForce RTX 3090 and Tesla V100 GPUs. 

\begin{table*}[t]
\caption{Results of ISIC17 dataset with 10\% and 20\% annotations. ``L/U" indicates the labeled and unlabeled samples.}
\label{tab: isic17}
\centering
\scalebox{0.9}{
\begin{tabular}{c|c|c|ccccccc}
\hline
\noalign{\smallskip}
L / U & Method & Venue & mIoU (\%) $\uparrow$ & Dice (\%) $\uparrow$ & Acc (\%) $\uparrow$ & Spe (\%) $\uparrow$ & Sen (\%) $\uparrow$ & 95HD $\downarrow$ & ASD $\downarrow$ \\ 
\noalign{\smallskip}
\hline
\noalign{\smallskip}
\multirow{10}{*}{\begin{tabular}[c]{@{}c@{}}300 / 1200 \\ (20 / 80\%)\end{tabular}}
& Mean Teacher\scalebox{0.7}{~\cite{tarvainen2017mean}} & NIPS'17 & 68.59$\pm$3.12 & 78.18$\pm$2.91 & 94.05$\pm$1.02 & 96.68$\pm$1.90 & 84.46$\pm$1.71 & 16.19$\pm$5.20 & 2.19$\pm$0.43 \\
& Fixmatch\scalebox{0.7}{~\cite{sohn2020fixmatch}} & NIPS'20 & 69.55$\pm$2.90 & 78.99$\pm$2.71 & 94.38$\pm$0.40 & 97.63$\pm$1.44 & 83.39$\pm$1.91 & 14.73$\pm$2.57 & 2.36$\pm$0.40 \\
& CPS\scalebox{0.7}{~\cite{chen2021semi}} & CVPR'21 & 68.91$\pm$2.92 & 78.33$\pm$3.04 & 94.36$\pm$0.38 & 97.41$\pm$0.64 & 82.86$\pm$3.33 & 15.41$\pm$3.38 & 2.75$\pm$1.02 \\
& SS-Net\scalebox{0.7}{~\cite{wu2022exploring}} & MICCAI'22 & 67.77$\pm$2.99 & 77.62$\pm$2.98 & 94.05$\pm$0.64 & 96.79$\pm$1.40 & 84.19$\pm$1.88 & 16.71$\pm$3.65 & 2.39$\pm$0.40 \\
& ACMT\scalebox{0.7}{~\cite{xu2023ambiguity}} & MedIA'23 & 69.53$\pm$5.04 & 78.94$\pm$4.68 & 93.80$\pm$2.33 & 96.32$\pm$3.71 & \textbf{86.04$\pm$2.83} & 16.77$\pm$10.47 & \textbf{1.93$\pm$0.43} \\
& BCP\scalebox{0.7}{~\cite{bai2023bidirectional}} & CVPR'23 & 70.83$\pm$3.68 & 80.27$\pm$2.90 & 94.74$\pm$0.54 & 97.97$\pm$1.47 & 84.79$\pm$4.03 & 11.56$\pm$2.11 & 2.20$\pm$0.58 \\
& IC\scalebox{0.7}{~\cite{huang2024exploring}} & TMI'24 & 68.97$\pm$4.17 & 77.83$\pm$4.51 & 94.33$\pm$0.56 & 97.75$\pm$0.61 & 80.96$\pm$5.69 & 13.88$\pm$2.51 & 2.98$\pm$1.29 \\
& CML\scalebox{0.7}{~\cite{wu2024cross}} & ACM MM'24 & 69.34$\pm$3.58 & 78.12$\pm$3.79 & 94.51$\pm$0.44 & \textbf{98.25$\pm$0.38} & 78.37$\pm$5.43 & 13.60$\pm$2.48 & 3.53$\pm$0.85 \\
& Semi-Mamba-UNet\scalebox{0.7}{~\cite{ma2024semi}} & Arxiv'24 & 70.34$\pm$2.06 & 79.24$\pm$2.23 & 94.70$\pm$0.06 & 97.65$\pm$0.05 & 80.96$\pm$2.13 & 11.87$\pm$1.93 & 2.67$\pm$0.32 \\
& \cellcolor{gray!20}\textbf{DCMamba (Ours)} & \cellcolor{gray!20}- & \cellcolor{gray!20}\textbf{74.23$\pm$1.24} & \cellcolor{gray!20}\textbf{82.58$\pm$1.11} & \cellcolor{gray!20}\textbf{95.57$\pm$0.03} & \cellcolor{gray!20}98.14$\pm$0.05 & \cellcolor{gray!20}84.17$\pm$1.31 & \cellcolor{gray!20}\textbf{9.43$\pm$0.68} & \cellcolor{gray!20}2.17$\pm$0.28 \\
\noalign{\smallskip}
\hline
\noalign{\smallskip}
\multirow{10}{*}{\begin{tabular}[c]{@{}c@{}}150 / 1350 \\ (10 / 90\%)\end{tabular}}
& Mean Teacher\scalebox{0.7}{~\cite{tarvainen2017mean}} & NIPS'17 & 63.06$\pm$4.86 & 73.10$\pm$4.51 & 92.13$\pm$2.43 & 95.74$\pm$3.86 & 79.87$\pm$3.19 & 20.45$\pm$6.60 & 3.86$\pm$1.09 \\
& Fixmatch\scalebox{0.7}{~\cite{sohn2020fixmatch}} & NIPS'20 & 65.71$\pm$4.68 & 75.14$\pm$4.93 & 93.46$\pm$0.60 & \textbf{98.29$\pm$0.94} & 76.46$\pm$7.46 & 15.79$\pm$2.23 & 3.71$\pm$1.42 \\
& CPS\scalebox{0.7}{~\cite{chen2021semi}} & CVPR'21 & 63.84$\pm$5.13 & 73.55$\pm$4.97 & 92.52$\pm$2.39 & 96.29$\pm$3.45 & 79.04$\pm$1.89 & 19.15$\pm$6.33 & 3.74$\pm$0.95 \\
& SS-Net\scalebox{0.7}{~\cite{wu2022exploring}} & MICCAI'22 & 62.11$\pm$5.10 & 72.04$\pm$4.92 & 91.76$\pm$2.98 & 95.80$\pm$4.50 & 78.19$\pm$3.74 & 20.48$\pm$6.92 & 4.02$\pm$0.85 \\
& ACMT\scalebox{0.7}{~\cite{xu2023ambiguity}} & MedIA'23 & 63.73$\pm$5.23 & 74.02$\pm$4.67 & 91.93$\pm$2.98 & 95.65$\pm$4.89 & 82.05$\pm$4.93 & 20.38$\pm$8.29 & 3.09$\pm$1.01 \\
& BCP\scalebox{0.7}{~\cite{bai2023bidirectional}} & CVPR'23 & 68.13$\pm$2.37 & 78.03$\pm$1.78 & 93.95$\pm$0.43 & 97.81$\pm$0.71 & 82.47$\pm$2.89 & 13.02$\pm$1.70 & 2.71$\pm$0.36 \\
& IC\scalebox{0.7}{~\cite{huang2024exploring}} & TMI'24 & 65.27$\pm$5.00 & 74.51$\pm$5.22 & 92.65$\pm$2.57 & 96.55$\pm$3.34 & 78.38$\pm$3.20 & 19.34$\pm$9.40 & 4.21$\pm$1.65 \\
& CML\scalebox{0.7}{~\cite{wu2024cross}} & ACM MM'24 & 63.67$\pm$4.48 & 72.96$\pm$4.18 & 92.84$\pm$0.77 & 97.58$\pm$1.38 & 73.38$\pm$8.29 & 16.60$\pm$3.07 & 4.46$\pm$1.78 \\
& Semi-Mamba-UNet\scalebox{0.7}{~\cite{ma2024semi}} & Arxiv'24 & 67.87$\pm$4.69 & 77.02$\pm$4.63 & 94.19$\pm$1.12 & 97.24$\pm$0.10 & 80.48$\pm$4.19 & 13.14$\pm$2.87 & 2.62$\pm$0.57 \\
& \cellcolor{gray!20}\textbf{DCMamba (Ours)} & \cellcolor{gray!20}- & \cellcolor{gray!20}\textbf{72.00$\pm$1.37} & \cellcolor{gray!20}\textbf{81.13$\pm$1.33} & \cellcolor{gray!20}\textbf{95.03$\pm$0.29} & \cellcolor{gray!20}98.07$\pm$0.06 & \cellcolor{gray!20}\textbf{83.23$\pm$2.46} & \cellcolor{gray!20}\textbf{10.56$\pm$0.98} & \cellcolor{gray!20}\textbf{2.41$\pm$0.43} \\
\noalign{\smallskip}
\hline
\end{tabular}}
\end{table*}

\begin{table*}[t]
\caption{Results of ISIC18 dataset with 10\% and 20\% annotations. ``L/U" indicates the labeled and unlabeled samples.}
\label{tab: isic18}
\centering
\scalebox{0.9}{
\begin{tabular}{c|c|c|ccccccc}
\hline
\noalign{\smallskip}
L / U & Method & Venue & mIoU (\%) $\uparrow$ & Dice (\%) $\uparrow$ & Acc (\%) $\uparrow$ & Spe (\%) $\uparrow$ & Sen (\%) $\uparrow$ & 95HD $\downarrow$ & ASD $\downarrow$ \\ 
\noalign{\smallskip}
\hline
\noalign{\smallskip}
\multirow{10}{*}{\begin{tabular}[c]{@{}c@{}}377 / 1509 \\ (20 / 80\%)\end{tabular}}
& Mean Teacher\scalebox{0.7}{~\cite{tarvainen2017mean}} & NIPS'17 & 71.72$\pm$2.85 & 81.02$\pm$2.49 & 91.45$\pm$1.22 & 96.50$\pm$1.88 & 81.69$\pm$6.11 & 14.39$\pm$2.21 & 3.19$\pm$1.26 \\
& Fixmatch\scalebox{0.7}{~\cite{sohn2020fixmatch}} & NIPS'20 & 73.62$\pm$1.92 & 82.55$\pm$1.67 & 92.08$\pm$0.93 & 97.09$\pm$1.33 & 82.49$\pm$4.55 & 12.90$\pm$1.36 & 2.98$\pm$1.02 \\
& CPS\scalebox{0.7}{~\cite{chen2021semi}} & CVPR'21 & 72.68$\pm$2.22 & 81.79$\pm$1.97 & 91.73$\pm$1.07 & 96.43$\pm$1.68 & 82.82$\pm$5.25 & 13.76$\pm$1.41 & 2.89$\pm$1.12 \\
& SS-Net\scalebox{0.7}{~\cite{wu2022exploring}} & MICCAI'22 & 71.54$\pm$3.73 & 80.74$\pm$3.50 & 91.34$\pm$1.49 & 96.46$\pm$1.66 & 81.77$\pm$6.55 & 14.71$\pm$2.10 & 3.27$\pm$1.42 \\
& ACMT\scalebox{0.7}{~\cite{xu2023ambiguity}} & MedIA'23 & 74.12$\pm$1.20 & 83.19$\pm$1.09 & 92.37$\pm$0.63 & 96.15$\pm$1.71 & 85.70$\pm$4.28 & 12.43$\pm$0.83 & 2.50$\pm$0.88 \\
& BCP\scalebox{0.7}{~\cite{bai2023bidirectional}} & CVPR'23 & 76.71$\pm$0.19 & 85.22$\pm$0.19 & 93.52$\pm$0.31 & 96.26$\pm$0.60 & 87.54$\pm$1.27 & 10.87$\pm$0.45 & 1.80$\pm$0.39 \\
& IC\scalebox{0.7}{~\cite{huang2024exploring}} & TMI'24 & 72.89$\pm$3.00 & 81.64$\pm$2.72 & 91.84$\pm$1.25 & \textbf{97.15$\pm$1.58} & 81.19$\pm$6.13 & 13.63$\pm$2.23 & 3.48$\pm$1.37 \\
& CML\scalebox{0.7}{~\cite{wu2024cross}} & ACM MM'24 & 74.70$\pm$1.41 & 83.27$\pm$1.35 & 92.53$\pm$0.77 & 96.31$\pm$1.07 & 84.18$\pm$3.62 & 12.28$\pm$1.46 & 2.71$\pm$0.89 \\
& Semi-Mamba-UNet\scalebox{0.7}{~\cite{ma2024semi}} & Arxiv'24 & 75.22$\pm$0.74 & 84.09$\pm$0.67 & 93.11$\pm$0.22 & 95.41$\pm$0.70 & 86.11$\pm$1.26 & 10.50$\pm$0.92 & 1.99$\pm$0.32 \\
& \cellcolor{gray!20}\textbf{DCMamba (Ours)} & \cellcolor{gray!20}- & \cellcolor{gray!20}\textbf{78.02$\pm$0.75} & \cellcolor{gray!20}\textbf{86.10$\pm$0.60} & \cellcolor{gray!20}\textbf{93.98$\pm$0.19} & \cellcolor{gray!20}95.85$\pm$0.33 & \cellcolor{gray!20}\textbf{87.99$\pm$1.27} & \cellcolor{gray!20}\textbf{8.51$\pm$0.48} & \cellcolor{gray!20}\textbf{1.67$\pm$0.20} \\
\noalign{\smallskip}
\hline
\noalign{\smallskip}
\multirow{10}{*}{\begin{tabular}[c]{@{}c@{}}187 / 1699 \\ (10 / 90\%)\end{tabular}}
& Mean Teacher\scalebox{0.7}{~\cite{tarvainen2017mean}} & NIPS'17 & 70.17$\pm$4.89 & 79.59$\pm$4.44 & 91.07$\pm$1.97 & 95.46$\pm$1.16 & 83.45$\pm$3.04 & 16.68$\pm$4.86 & 2.78$\pm$0.93 \\
& Fixmatch\scalebox{0.7}{~\cite{sohn2020fixmatch}} & NIPS'20 & 71.94$\pm$3.24 & 81.16$\pm$2.83 & 91.79$\pm$1.16 & 96.79$\pm$0.81 & 83.23$\pm$1.96 & 14.50$\pm$3.41 & 2.73$\pm$0.76 \\
& CPS\scalebox{0.7}{~\cite{chen2021semi}} & CVPR'21 & 72.18$\pm$1.93 & 81.42$\pm$1.68 & 92.05$\pm$0.70 & 96.23$\pm$0.97 & 84.37$\pm$2.78 & 14.32$\pm$1.29 & 2.58$\pm$0.84 \\
& SS-Net\scalebox{0.7}{~\cite{wu2022exploring}} & MICCAI'22 & 69.12$\pm$4.91 & 78.53$\pm$4.48 & 90.90$\pm$1.44 & 96.58$\pm$1.10 & 80.17$\pm$3.83 & 16.97$\pm$3.45 & 3.76$\pm$1.30 \\
& ACMT\scalebox{0.7}{~\cite{xu2023ambiguity}} & MedIA'23 & 72.64$\pm$2.12 & 81.83$\pm$1.74 & 91.99$\pm$0.74 & 96.22$\pm$1.86 & 85.12$\pm$2.46 & 14.14$\pm$2.05 & 2.58$\pm$0.60 \\
& BCP\scalebox{0.7}{~\cite{bai2023bidirectional}} & CVPR'23 & 73.69$\pm$3.10 & 82.72$\pm$2.32 & 92.54$\pm$0.92 & 95.61$\pm$2.21 & 87.17$\pm$3.71 & 12.69$\pm$2.02 & 2.05$\pm$0.73 \\
& IC\scalebox{0.7}{~\cite{huang2024exploring}} & TMI'24 & 71.91$\pm$2.53 & 80.71$\pm$2.23 & 91.53$\pm$1.10 & 96.29$\pm$1.60 & 82.39$\pm$3.56 & 14.76$\pm$2.10 & 3.31$\pm$1.09 \\
& CML\scalebox{0.7}{~\cite{wu2024cross}} & ACM MM'24 & 71.87$\pm$3.21 & 80.55$\pm$3.10 & 91.71$\pm$1.17 & \textbf{97.37$\pm$0.40} & 79.69$\pm$4.18 & 13.93$\pm$1.84 & 3.69$\pm$0.93 \\
& Semi-Mamba-UNet\scalebox{0.7}{~\cite{ma2024semi}} & Arxiv'24 & 72.74$\pm$0.02 & 81.74$\pm$0.02 & 92.32$\pm$0.01 & 94.86$\pm$0.03 & 85.11$\pm$0.04 & 13.19$\pm$2.49 & 2.11$\pm$0.79 \\
& \cellcolor{gray!20}\textbf{DCMamba (Ours)} & \cellcolor{gray!20}- & \cellcolor{gray!20}\textbf{77.27$\pm$1.44} & \cellcolor{gray!20}\textbf{85.45$\pm$1.15} & \cellcolor{gray!20}\textbf{93.73$\pm$0.57} & \cellcolor{gray!20}96.27$\pm$1.22 & \cellcolor{gray!20}\textbf{87.59$\pm$1.55} & \cellcolor{gray!20}\textbf{9.73$\pm$1.21} & \cellcolor{gray!20}\textbf{1.79$\pm$0.29} \\
\noalign{\smallskip}
\hline
\end{tabular}}
\end{table*}

\begin{figure*}[t]
   \centering
   \includegraphics[width=0.95\linewidth]{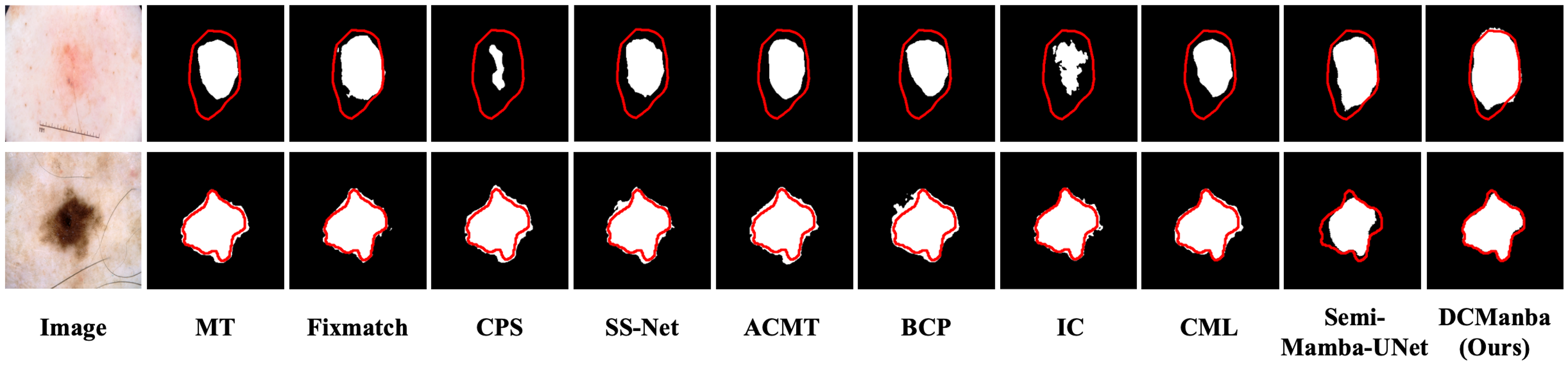}
   \caption{The segmentation results on the ISIC18 dataset, where the red curves denote corresponding ground truth labels. }
   \label{fig: isic18}
\end{figure*}

In our experiments, we adopted horizontal and vertical, diagonal and anti-diagonal scanning directions, which exhibited notable differences, as key directions for collaborative networks to enhance diversity.
We randomly select one of the networks for inference and discard the other. For a fair comparison, we use the results of the network with horizontal and vertical scans. 

\begin{table*}[t]
\caption{Results of Kvasir-SEG dataset with 10\% and 20\% annotations. ``L/U" indicates the labeled and unlabeled samples.}
\label{tab: kvasir}
\centering
\scalebox{0.9}{
\begin{tabular}{c|c|c|ccccccc}
\hline
\noalign{\smallskip}
L / U & Method & Venue & mIoU (\%) $\uparrow$ & Dice (\%) $\uparrow$ & Acc (\%) $\uparrow$ & Spe (\%) $\uparrow$ & Sen (\%) $\uparrow$ & 95HD $\downarrow$ & ASD $\downarrow$ \\ 
\noalign{\smallskip}
\hline
\noalign{\smallskip}
\multirow{10}{*}{\begin{tabular}[c]{@{}c@{}}160 / 640 \\ (20 / 80\%)\end{tabular}}
& Mean Teacher\scalebox{0.7}{~\cite{tarvainen2017mean}} & NIPS'17 & 62.41$\pm$3.71 & 71.78$\pm$3.58 & 93.48$\pm$0.70 & 98.41$\pm$0.25 & 71.65$\pm$3.47 & 26.56$\pm$5.04 & 5.93$\pm$1.78 \\
& Fixmatch\scalebox{0.7}{~\cite{sohn2020fixmatch}} & NIPS'20 & 64.77$\pm$2.66 & 73.68$\pm$2.54 & 93.85$\pm$0.52 & 98.80$\pm$0.15 & 72.10$\pm$2.59 & 22.89$\pm$3.22 & 5.51$\pm$0.99 \\
& CPS\scalebox{0.7}{~\cite{chen2021semi}} & CVPR'21 & 63.73$\pm$2.87 & 72.77$\pm$2.95 & 93.68$\pm$0.54 & 98.34$\pm$0.12 & 73.25$\pm$3.39 & 24.93$\pm$2.14 & 5.67$\pm$1.31 \\
& SS-Net\scalebox{0.7}{~\cite{wu2022exploring}} & MICCAI'22 & 63.10$\pm$2.91 & 72.65$\pm$2.82 & 93.51$\pm$0.57 & 97.63$\pm$0.22 & 75.54$\pm$2.69 & 29.38$\pm$2.75 & 5.32$\pm$1.12 \\
& ACMT\scalebox{0.7}{~\cite{xu2023ambiguity}} & MedIA'23 & 65.66$\pm$2.13 & 74.73$\pm$2.07 & 93.94$\pm$0.44 & 98.04$\pm$0.34 & 76.83$\pm$2.07 & 24.06$\pm$3.00 & 4.30$\pm$0.49 \\
& BCP\scalebox{0.7}{~\cite{bai2023bidirectional}} & CVPR'23 & 67.04$\pm$1.41 & 75.88$\pm$1.17 & 93.99$\pm$0.24 & \textbf{99.14$\pm$0.22} & 72.20$\pm$2.01 & 19.41$\pm$0.91 & 4.86$\pm$0.68 \\
& IC\scalebox{0.7}{~\cite{huang2024exploring}} & TMI'24 & 64.85$\pm$2.74 & 73.32$\pm$2.72 & 93.99$\pm$0.67 & 98.79$\pm$0.11 & 72.58$\pm$2.87 & 22.88$\pm$2.03 & 5.93$\pm$0.81 \\
& CML\scalebox{0.7}{~\cite{wu2024cross}} & ACM MM'24 & 72.60$\pm$0.79 & 81.02$\pm$0.62 & 94.91$\pm$0.23 & 98.73$\pm$0.28 & 79.94$\pm$1.42 & 17.78$\pm$0.78 & 3.88$\pm$0.49 \\
& Semi-Mamba-UNet\scalebox{0.7}{~\cite{ma2024semi}} & Arxiv'24 & 67.61$\pm$7.00 & 76.90$\pm$6.03 & 95.79$\pm$1.23 & 97.91$\pm$0.57 & 79.42$\pm$4.58 & 21.65$\pm$5.90 & 4.68$\pm$1.96 \\
& \cellcolor{gray!20}\textbf{DCMamba (Ours)} & \cellcolor{gray!20}- & \cellcolor{gray!20}\textbf{78.70$\pm$0.21} & \cellcolor{gray!20}\textbf{86.31$\pm$0.26} & \cellcolor{gray!20}\textbf{96.32$\pm$0.05} & \cellcolor{gray!20}98.50$\pm$0.03 & \cellcolor{gray!20}\textbf{87.07$\pm$0.22} & \cellcolor{gray!20}\textbf{12.19$\pm$0.22} & \cellcolor{gray!20}\textbf{2.10$\pm$0.16} \\
\noalign{\smallskip}
\hline
\noalign{\smallskip} 
\multirow{10}{*}{\begin{tabular}[c]{@{}c@{}}80 / 720 \\ (10 / 90\%)\end{tabular}}
& Mean Teacher\scalebox{0.7}{~\cite{tarvainen2017mean}} & NIPS'17 & 51.87$\pm$3.42 & 61.70$\pm$3.79 & 91.20$\pm$0.72 & 96.92$\pm$0.49 & 64.65$\pm$5.07 & 35.85$\pm$3.27 & 8.42$\pm$1.01 \\
& Fixmatch\scalebox{0.7}{~\cite{sohn2020fixmatch}} & NIPS'20 & 57.84$\pm$1.47 & 67.57$\pm$1.77 & 92.36$\pm$0.23 & 97.44$\pm$0.59 & 70.13$\pm$3.60 & 31.97$\pm$2.15 & 6.57$\pm$1.47 \\
& CPS\scalebox{0.7}{~\cite{chen2021semi}} & CVPR'21 & 52.06$\pm$2.78 & 61.61$\pm$2.92 & 91.47$\pm$0.56 & 97.67$\pm$0.10 & 62.74$\pm$3.99 & 33.73$\pm$2.71 & 9.43$\pm$1.05 \\
& SS-Net\scalebox{0.7}{~\cite{wu2022exploring}} & MICCAI'22 & 52.44$\pm$2.73 & 62.46$\pm$3.02 & 91.10$\pm$0.74 & 96.36$\pm$0.92 & 67.21$\pm$3.17 & 37.93$\pm$4.21 & 8.18$\pm$0.97 \\
& ACMT\scalebox{0.7}{~\cite{xu2023ambiguity}} & MedIA'23 & 55.68$\pm$1.71 & 65.63$\pm$2.11 & 91.71$\pm$0.82 & 96.81$\pm$1.29 & 69.70$\pm$3.51 & 32.17$\pm$4.93 & 6.43$\pm$0.68 \\
& BCP\scalebox{0.7}{~\cite{bai2023bidirectional}} & CVPR'23 & 62.73$\pm$1.41 & 71.59$\pm$1.38 & 93.24$\pm$0.18 & \textbf{99.33$\pm$0.18} & 66.96$\pm$1.91 & 21.74$\pm$0.99 & 6.51$\pm$0.51 \\
& IC\scalebox{0.7}{~\cite{huang2024exploring}} & TMI'24 & 55.41$\pm$2.28 & 64.28$\pm$2.30 & 92.24$\pm$0.56 & 98.66$\pm$0.28 & 63.24$\pm$2.09 & 27.58$\pm$2.51 & 7.82$\pm$1.63 \\
& CML\scalebox{0.7}{~\cite{wu2024cross}} & ACM MM'24 & 69.74$\pm$1.60 & 78.37$\pm$1.51 & 94.40$\pm$0.18 & 98.72$\pm$0.32 & 77.16$\pm$1.47 & 19.24$\pm$2.34 & 4.41$\pm$0.38 \\
& Semi-Mamba-UNet\scalebox{0.7}{~\cite{ma2024semi}} & Arxiv'24 & 57.35$\pm$10.04 & 67.46$\pm$8.83 & 91.92$\pm$2.30 & 96.76$\pm$1.66 & 72.22$\pm$5.62 & 31.86$\pm$10.90 & 7.05$\pm$2.77 \\
& \cellcolor{gray!20}\textbf{DCMamba (Ours)} & \cellcolor{gray!20}- & \cellcolor{gray!20}\textbf{74.54$\pm$0.26} & \cellcolor{gray!20}\textbf{83.02$\pm$0.27} & \cellcolor{gray!20}\textbf{95.58$\pm$0.07} & \cellcolor{gray!20}98.26$\pm$0.08 & \cellcolor{gray!20}\textbf{84.40$\pm$0.53} & \cellcolor{gray!20}\textbf{15.89$\pm$0.40} & \cellcolor{gray!20}\textbf{2.71$\pm$0.37} \\
\noalign{\smallskip}
\hline
\end{tabular}}
\end{table*}

\begin{figure*}[t]
   \centering
   \includegraphics[width=0.96\linewidth]{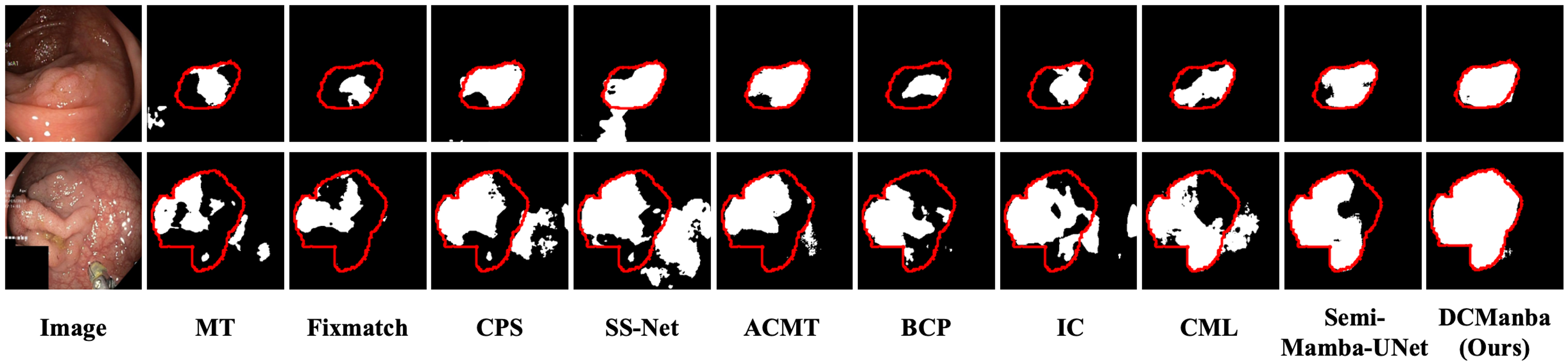}
   \caption{The segmentation results on the Kvasir dataset, where the red curves denote corresponding ground truth labels. }
   \label{fig: kvasir}
\end{figure*}

\subsection{Results on 2D Tasks}
We compare our proposed DCMamba with various baselines and SOTA SSL algorithms, including Mean Teacher~\cite{tarvainen2017mean}, Fixmatch~\cite{sohn2020fixmatch}, CPS~\cite{chen2021semi}, SS-Net~\cite{wu2022exploring}, ACMT~\cite{xu2023ambiguity}, BCP~\cite{bai2023bidirectional}, IC~\cite{huang2024exploring}, CML~\cite{wu2024cross}, and Semi-Mamba-UNet~\cite{ma2024semi}. 
Among them, FixMatch~\cite{sohn2020fixmatch}, BCP~\cite{bai2023bidirectional} and CML~\cite{wu2024cross} are semi-supervised methods that utilize strong enhancement.
To mitigate the potential bias introduced by varying data splits, we implement the robust evaluation by running five trials with different random samples of unlabeled data. The results are presented in the format of $mean \pm std$.

\subsubsection{Evaluation on Synapse Dataset}
The results on the Synapse dataset are shown in Table~\ref{tab: synapse}. We report the Dice Similarity Coefficient (Dice) as an evaluation metric. As there’s no validation set in the experiments, we report the results at the end of 20,000 training iterations for all methods for fair comparison. 
It can be observed that SSM-based methods outperform other methods on both 20\% and 10\% labeled data scenarios. In addition, our methods beat the latest Semi-Mamba-UNet~\cite{ma2024semi}, with 4.52\% to 6.69\% higher Dice scores on average, which proves the effectiveness of our method for semi-supervised learning. The improvement in Dice scores is evident across most organ classes, particularly for the Right Kidney, Left Kidney, Spleen, and Stomach, where DCMamba's directional diversity and collaborative learning strategies enhance its ability to capture the cross-organ relationships and generalize across different organ types.

\subsubsection{Evaluation on ISIC17 Dataset}
The evaluations on Mean Intersection over Union (mIoU), Dice Similarity Coefficient (Dice), Accuracy (Acc), Specificity (Spe), and Sensitivity (Sen) metrics. These results do not conduct any post-processing for fair comparison. 
The experimental results are presented in Table~\ref{tab: isic17}. 
It is clear that our method outperforms other SOTA semi-supervised methods in mIoU, Dice, Acc, and 95HD. Our DCMamba achieves the best performance among all competitors, surpassing the top SSL method BCP~\cite{bai2023bidirectional} by 3.40\% (74.23\% vs. 70.83\%) on the mIoU metric. Additionally, our method outperforms the latest SSM-based method Semi-Mamba-UNet by 3.89\% (74.23\% vs. 70.34\%).

\subsubsection{Evaluation on ISIC18 Dataset}
We evaluate the effectiveness of our method on ISIC18 dataset and present the results in Table~\ref{tab: isic18}.
We have observed that some methods, \eg, IC~\cite{huang2024exploring} and CML~\cite{wu2024cross},  performed well in Specificity but mediocre in other indicators, which may be because these methods tend to over-predict the negative class, that is, the background pixel area, and it also causes the model to perform poorly in identifying and segmenting the target area. Our methods achieve significant improvements compared to other SOTA semi-supervised methods, \eg, DCMamba outperforms the SOTA method BCP~\cite{bai2023bidirectional} by 1.31\% on mIoU with 20\% labeled data. Also, DCMamba achieves a performance improvement of up to 3.58\% (77.27\% vs. 73.69\%) over the second-best result under the setting of 10\% labeling ratio.

\begin{table*}[t]
\caption{Results of 3D Synapse dataset with 20\% annotations. ``L/U" indicates the labeled and unlabeled samples.\\ * means all the pixels are predicted as background or another region.}
\label{tab: synapse3d}
\centering
\scalebox{0.85}{
\begin{tabular}{c|c|c|cc|cccccccc}
\hline
\noalign{\smallskip}
\multirow{2.5}{*}{L / U} & \multirow{2.5}{*}{Method} & \multirow{2.5}{*}{Venue} & \multirow{2.5}{*}{\begin{tabular}[c]{@{}c@{}}Average \\ Dice (\%) $\uparrow$\end{tabular}} & \multirow{2.5}{*}{\begin{tabular}[c]{@{}c@{}}Average \\ ASD $\downarrow$\end{tabular}} & \multicolumn{8}{c}{Average Dice of Each Class (\%) $\uparrow$} \\ 
\noalign{\smallskip}
\cline{6-13}
\noalign{\smallskip}
& & & & & Arota & Gallbladder & R.Kedney & L.Kedney & Liver & Pancreas & Spleen & Stomach  \\ 
\noalign{\smallskip}
\hline
\noalign{\smallskip}
\multirow{9}{*}{\begin{tabular}[c]{@{}c@{}}4 / 14 \\ (20 / 80\%)\end{tabular}}
& Mean Teacher\scalebox{0.7}{~\cite{tarvainen2017mean}} & NIPS'17 & 45.62 & 43.12 & 56.71 & 27.06 & 59.09 & 65.66 & 66.51 & 22.49 & 26.00 & 41.47 \\
& Fixmatch\scalebox{0.7}{~\cite{sohn2020fixmatch}} & NIPS'20 & 51.21 & 35.65 & 82.20 & 11.30 & 66.18 & 62.21 & 66.64 & 32.37 & 77.95 & 10.81 \\
& CPS\scalebox{0.7}{~\cite{chen2021semi}} & CVPR'21 & 46.24 & 15.87 & 72.01 & 9.33 & 56.87 & 64.86 & 83.29 & 1.99 & 60.53 & 21.01 \\
& SS-Net\scalebox{0.7}{~\cite{wu2022exploring}} & MICCAI'22 & 41.49 & 40.25 & 66.50 & 23.50 & 23.51 & 37.66 & 77.11 & 8.01 & 62.88 & 32.78 \\
& ACMT\scalebox{0.7}{~\cite{xu2023ambiguity}} & MedIA'23 & 54.91 & 25.58 & 82.95 & 25.14 & 72.51 & 71.59 & 86.86 & 26.45 & 37.20 & 36.59 \\
& BCP\scalebox{0.7}{~\cite{bai2023bidirectional}} & CVPR'23 & 59.83 & 15.57 & 80.31 & 24.96 & 65.09 & 79.72 & 83.46 & 29.14 & \textbf{81.07} & 34.85 \\
& IC\scalebox{0.7}{~\cite{huang2024exploring}} & TMI'24 & 52.30 & 22.20 & 61.36 & 12.64 & 71.46 & 62.55 & 90.34 & 4.63 & 84.75 & 30.65 \\
& CML\scalebox{0.7}{~\cite{wu2024cross}} & ACM MM'24 & 55.92 & 21.25 & \textbf{85.67} & 0* & 52.38 & \textbf{82.75} & \textbf{92.97} & 20.65 & 64.70 & 48.22 \\
& \cellcolor{gray!20}\textbf{DCMamba (Ours)} & \cellcolor{gray!20}- & \cellcolor{gray!20}\textbf{66.05} & \cellcolor{gray!20}\textbf{11.17} & \cellcolor{gray!20}83.44 & \cellcolor{gray!20}\textbf{34.65} & \cellcolor{gray!20}\textbf{75.28} & \cellcolor{gray!20}79.33 & \cellcolor{gray!20}90.68 & \cellcolor{gray!20}\textbf{41.72} & \cellcolor{gray!20}66.61 & \cellcolor{gray!20}\textbf{56.68} \\
\noalign{\smallskip}
\hline
\end{tabular}}
\end{table*}

\begin{table*}[t]
\caption{Results of FLARE22 dataset with 10\% annotations. ``L/U" indicates the labeled and unlabeled samples.}
\label{tab: flare}
\centering
\scalebox{0.8}{
\begin{tabular}{c|c|ccccccccccccc|cc}
\toprule
\multirow{2.5}{*}{L / U} & \multirow{2.5}{*}{Method} & \multicolumn{13}{c|}{Average Dice of Each Class (\%) $\uparrow$} & \multirow{2.5}{*}{\begin{tabular}[c]{@{}c@{}}Average \\ Dice (\%) $\uparrow$\end{tabular}} & \multirow{2.5}{*}{\begin{tabular}[c]{@{}c@{}}Average \\ Jaccard (\%) $\uparrow$\end{tabular}} \\ 
\noalign{\smallskip}
\cline{3-15}
\noalign{\smallskip}
& & Liv & Spl & Sto & L.kid & R.kid & Aor & Pan & IVC & Duo & Gal & Eso & RAG & LAG \\
\noalign{\smallskip}
\hline
\noalign{\smallskip}
\multirow{11}{*}{\begin{tabular}[c]{@{}c@{}}42 / 378 \\ (10 / 90\%)\end{tabular}}
& DAN~\cite{zhang2017deep} & 95.85 & 84.15 & 67.29 & 92.81 & 91.96 & 91.35 & 63.12 & 79.33 & 66.48 & 77.29 & 67.82 & 50.41 & 48.41 & 75.10 $\pm$ 0.69 & 63.83 $\pm$ 0.23 \\
& Mean Teacher~\cite{tarvainen2017mean} & 96.49 & 91.54 & 74.64 & 93.78 & 92.77 & 92.17 & 69.23 & 82.71 & 66.68 & 73.49 & 70.66 & 61.88 & 41.26 & 77.49 $\pm$ 0.48 & 66.45 $\pm$ 0.39 \\
& UA-MT~\cite{yu2019uncertainty} & 96.42 & 91.98 & 79.91 & 92.74 & 92.83 & 92.33 & 71.43 & 83.10 & 67.72 & 77.26 & 72.41 & 64.04 & 46.18 & 79.10 $\pm$ 0.38 & 68.21 $\pm$ 0.48 \\
& SASSNet~\cite{li2020shape} & 96.21 & 90.40 & 67.12 & 94.00 & 92.85 & 91.61 & 67.89 & 79.59 & 65.47 & 71.59 & 71.44 & 52.07 & 57.83 & 76.77 $\pm$ 0.30 & 65.89 $\pm$ 0.44 \\
& DTC~\cite{luo2021semi} & 96.63 & 92.91 & 72.76 & 92.68 & 92.40 & 91.87 & 66.82 & 81.47 & 65.76 & 78.38 & 69.39 & 59.74 & 59.10 & 78.45 $\pm$ 0.82 & 67.05 $\pm$ 1.02 \\
& CPS~\cite{chen2021semi} & 96.62 & 92.16 & 77.02 & 92.70 & 92.71 & 92.25 & 69.39 & 81.91 & 65.94 & 75.12 & 72.78 & 63.56 & 58.96 & 79.32 $\pm$ 0.46 & 68.14 $\pm$ 0.61 \\
& CLD~\cite{lin2022calibrating} & 94.63 & 89.74 & 73.20 & 91.76 & 92.97 & 91.61 & 70.27 & 83.12 & 68.13 & 84.15 & 72.69 & 67.89 & 55.27 & 79.65 $\pm$ 0.17 & 68.22 $\pm$ 0.49 \\
& DHC~\cite{wang2023dhc} & 93.17 & 90.64 & 80.56 & 93.13 & 92.89 & 91.38 & 72.22 & 83.75 & 69.73 & 82.47 & 73.25 & 67.12 & 56.19 & 80.50 $\pm$ 0.43 & 69.38 $\pm$ 0.63 \\
& MagicNet~\cite{chen2023magicnet} & 97.04 & 88.04 & 81.51 & 92.18 & 92.95 & 91.75 & 71.15 & 81.01 & 69.61 & 84.36 & 77.07 & 63.34 & 60.33 & 80.79 $\pm$ 0.75 & 70.23 $\pm$ 0.96 \\
& GuidedNet~\cite{zhao2024guidednet} & 96.77 & 93.48 & 83.19 & 94.51 & \textbf{93.48} & 92.95 & 75.97 & 84.63 & 71.92 & \textbf{85.87} & \textbf{75.74} & 71.10 & 67.77 & 83.64 $\pm$ 0.42 & 73.08 $\pm$ 0.38 \\
& \cellcolor{gray!20}\textbf{DCMamba (Ours)} & \cellcolor{gray!20}\textbf{97.60} & \cellcolor{gray!20}\textbf{95.20} & \cellcolor{gray!20}\textbf{89.89} & \cellcolor{gray!20}\textbf{95.62} & \cellcolor{gray!20}92.88 & \cellcolor{gray!20}\textbf{93.78} & \cellcolor{gray!20}\textbf{80.75} & \cellcolor{gray!20}\textbf{85.53} & \cellcolor{gray!20}\textbf{73.86} & \cellcolor{gray!20}63.61 & \cellcolor{gray!20}75.47 & \cellcolor{gray!20}\textbf{73.99} & \cellcolor{gray!20}\textbf{76.17} & \cellcolor{gray!20}\textbf{84.18 $\pm$ 0.11} & \cellcolor{gray!20}\textbf{75.87 $\pm$ 0.13} \\
\bottomrule
\end{tabular}}
\end{table*}

\begin{table}[t]
\caption{Results of AbdomenCT-1K dataset with 20\% annotations. ``L/U" indicates the labeled and unlabeled samples.}
\label{tab: abdomen1k}
\centering
\scalebox{0.6}{
\begin{tabular}{c|c|c|cc|cccc}
\hline
\noalign{\smallskip}
\multirow{2.5}{*}{L / U} & \multirow{2.5}{*}{Method} & \multirow{2.5}{*}{Venue} & \multirow{2.5}{*}{\begin{tabular}[c]{@{}c@{}}Average \\ Dice (\%) $\uparrow$\end{tabular}} & \multirow{2.5}{*}{\begin{tabular}[c]{@{}c@{}}Average \\ ASD $\downarrow$\end{tabular}} & \multicolumn{4}{c}{Average Dice of Each Class (\%) $\uparrow$} \\ 
\noalign{\smallskip}
\cline{6-9}
\noalign{\smallskip}
& & & & & Liver & Kidney & Spleen & Pancreas \\ 
\noalign{\smallskip}
\hline
\noalign{\smallskip}
\multirow{9}{*}{\begin{tabular}[c]{@{}c@{}}180 / 720 \\ (20 / 80\%)\end{tabular}}
& Fixmatch\scalebox{0.7}{~\cite{sohn2020fixmatch}} & NIPS'20 & 89.36 & 2.17 & 96.32 & \textbf{94.72} & 93.59 & 72.81 \\
& CPS\scalebox{0.7}{~\cite{chen2021semi}} & CVPR'21 & 89.68 & 2.23 & 96.42 & 94.48 & 94.19 & 73.62 \\
& SS-Net\scalebox{0.7}{~\cite{wu2022exploring}} & MICCAI'22 & 85.69 & 5.24 & 93.09 & 91.40 & 90.48 & 67.78 \\
& ACMT\scalebox{0.7}{~\cite{xu2023ambiguity}} & MedIA'23 & 89.56 & 1.83 & 96.62 & 94.33 & 93.44 & 73.84 \\
& BCP\scalebox{0.7}{~\cite{bai2023bidirectional}} & CVPR'23 & 86.41 & 6.02 & 95.72 & 88.30 & 92.10 & 69.52 \\
& MagicNet\scalebox{0.7}{~\cite{chen2023magicnet}} & CVPR'23 & 90.01 & 2.56 & 96.13 & 94.09 & 95.12 & 74.71 \\
& IC\scalebox{0.7}{~\cite{huang2024exploring}} & TMI'24 & 86.71 & 3.48 & 96.41 & 93.64 & 91.50 & 65.31 \\
& CML\scalebox{0.7}{~\cite{wu2024cross}} & ACM MM'24 & 89.66 & 4.15 & 96.39 & 93.84 & 93.86 & 74.57 \\
& \cellcolor{gray!20}\textbf{DCMamba (Ours)} & \cellcolor{gray!20}- & \cellcolor{gray!20}\textbf{90.21} & \cellcolor{gray!20}\textbf{1.19} & \cellcolor{gray!20}\textbf{96.81} & \cellcolor{gray!20}94.45 & \cellcolor{gray!20}\textbf{95.78} & \cellcolor{gray!20}\textbf{74.79} \\
\noalign{\smallskip}
\hline
\end{tabular}}
\end{table}

\subsubsection{Evaluation on Kvasir-SEG Dataset}
As presented in Table~\ref{tab: kvasir}, we observe that our DCMamba significantly outperforms other methods, achieving a SOTA performance of 8.70\% on mIoU with 20\% labeled data. Also, DCMamba shows marked superiority in low annotation settings in both segmentation accuracy and boundary precision, suggesting that its diversity-enhanced framework ensures stable convergence and consistent improvements over competitors.

Our approach employs the SS2D module to scan image features along multiple directions — horizontal, vertical, diagonal, and anti-diagonal — by sequentially accumulating information along the scanning path. This mechanism enables the network to have a strong perception of information in certain directions and the ability to capture spatial context progressively. Although ISIC17, ISIC18, and Kvasir-SEG datasets do not exhibit standardized anatomical orientation like CT or MR scans, this does not imply the absence of meaningful spatial patterns. Lesions and polyps often exhibit local texture patterns with certain directionality, for example, radial streaks around skin lesions or uneven lighting across the background. These subtle variations can be effectively captured by different scan directions to induce differences in feature aggregation. Our method achieved consistent and significant improvements on these datasets, further demonstrating the effectiveness and generalizability across different medical imaging modalities.

\subsubsection{Visualization}
The segmentation result examples on Synapse, ISIC18, and Kvasir-SEG are shown in Figure~\ref{fig: synapse}, \ref{fig: isic18}, and \ref{fig: kvasir}. Through the visualization of the results, we could clearly observe that our method demonstrates outstanding performance in capturing the holistic shape structure. Furthermore, our DCMamba also exhibits sensitivity to subtle details, particularly in the edge region of the organs and lesions.

Across the four datasets, the performance differences across datasets could be attributed to varying task complexities. The Synapse dataset requires segmenting multiple organs with different shapes and textures. Despite this, DCMamba shows robust improvement with a Dice score of over 60\%. For ISIC17, ISIC18, and Kvasir datasets, which are single-object segmentation tasks with slightly lower difficulty than Synapse, DCMamba achieves Dice scores above 80\% and ASD scores below 3 with only 10\% annotations.

\begin{table*}[t]
\caption{Effectiveness of each component in DCMamba on four datasets with 20\% labeled data.}
\label{tab: component}
\centering
\scalebox{0.9}{
\begin{tabular}{ccc|cc|cc|cc|cc}
\hline
\noalign{\smallskip}
\multicolumn{3}{c|}{Key Components} & \multicolumn{2}{c|}{Synapse} & \multicolumn{2}{c|}{ISIC18} & \multicolumn{2}{c|}{ISIC17} & \multicolumn{2}{c}{Kvasir} \\
\noalign{\smallskip}
diverse-augment & diverse-scan & diverse-feature & Dice (\%) $\uparrow$ & ASD $\downarrow$ & Dice (\%) $\uparrow$ & ASD $\downarrow$ & Dice (\%) $\uparrow$ & ASD $\downarrow$ & Dice (\%) $\uparrow$ & ASD $\downarrow$ \\ 
\noalign{\smallskip}
\hline
\noalign{\smallskip}
\ding{56} & \ding{56} & \ding{56} & 58.00 & 22.55 & 84.01 & 1.71 & 72.06 & 3.31 & 83.01 & 2.62 \\
\noalign{\smallskip}
\hline
\noalign{\smallskip}
\ding{52} & \ding{56} & \ding{56} & 61.07 & 21.80 & 85.37 & 2.08 & 78.99 & 2.63 & 84.90 & 2.14 \\
\ding{56} & \ding{52} & \ding{56} & 61.62 & 19.81 & 84.78 & 1.87 & 76.67 & 2.70 & 84.86 & 2.04 \\
\ding{56} & \ding{56} & \ding{52} & 59.09 & 26.41 & 84.59 & 2.05 & 76.93 & 2.27 & 83.89 & 2.66 \\
\ding{52} & \ding{52} & \ding{56} & 64.60 & 12.52 & 85.57 & 1.99 & 80.52 & 2.12 & 86.01 & 2.32 \\
\ding{56} & \ding{52} & \ding{52} & 63.39 & 20.76 & 85.02 & 1.99 & 77.53 & 2.24 & 85.03 & \textbf{1.91} \\
\noalign{\smallskip}
\hline
\noalign{\smallskip}
\ding{52} & \ding{52} & \ding{52} & \textbf{67.63} & \textbf{10.70} & \textbf{86.42} & \textbf{1.66} & \textbf{81.05} & \textbf{2.11} & \textbf{86.44} & 2.04 \\
\noalign{\smallskip}
\hline
\end{tabular}}
\end{table*}

\subsection{Results on 3D Tasks}
\subsubsection{Evaluation on Synapse Dataset}
As shown in Table~\ref{tab: synapse3d}, we compare the performance of our DCMamba framework with several recent semi-supervised learning methods on the 3D Synapse dataset, using 20\% labeled data. Our results reveal that DCMamba outperforms existing methods across both Average Dice and Average ASD metrics, demonstrating the effectiveness of our approach in 3D medical image segmentation. Our DCMamba achieves an Average Dice of 66.05\%, significantly outperforming the second-best method, CML~\cite{wu2024cross} (55.92\%). Notably, some methods (IC~\cite{huang2024exploring} and CML~\cite{wu2024cross}) struggle to detect certain organs, such as the gallbladder. In contrast, our approach leverages a diverse-scan collaborative mechanism that allows for a comprehensive integration of contextual information from multiple perspectives. This enables the model to capture the spatial relationships between organs in 3D space, ensuring that even small or ambiguous structures are detected. The experimental results demonstrate the potential for adapting our method to 3D medical image segmentation tasks.

\subsubsection{Evaluation on AbdomenCT-1K Dataset}
We present the results on the AbdomenCT-1K dataset using 20\% labeled annotations in Table \ref{tab: abdomen1k}. Our method, DCMamba, achieves the highest average Dice score of 90.21\% and the lowest ASD of 1.19, demonstrating its superior segmentation accuracy and boundary precision. For class-wise performance, DCMamba achieves the highest Dice scores for Liver, Spleen, and Pancreas, while maintaining competitive performance for Kidney.

\subsubsection{Evaluation on FLARE22 Dataset}
The experimental results are presented in Table~\ref{tab: flare}. It could be observed that with 10\% annotations, our framework achieves the best performance across both Dice and Jaccard metrics. Specifically, DCMamba attains an average Dice score of 84.18\%, significantly outperforming the second-best method, GuidedNet~\cite{zhao2024guidednet} (83.64\%). Furthermore, DCMamba achieves the highest Dice scores in most organ segmentation tasks, including liver, spleen, stomach, left kidney, aorta, pancreas, IVC, duodenum, RAG, and LAG, demonstrating its superior generalization ability. While its performance on the gallbladder was lower than CNN-based methods, this may be attributed to its selective scanning mechanism, which prioritizes long-range dependencies over local details. Nevertheless, these results highlight the robustness and effectiveness of DCMamba in semi-supervised multi-organ segmentation on the FLARE22 dataset.

\subsection{Ablation Studies}
\subsubsection{Effectiveness of Key Components}
We conduct ablation studies to evaluate the effectiveness of each component in our framework, and all experiments are done on the Synapse, ISIC17, and Kvasir datasets.
Table~\ref{tab: component} presents the result of removing some structures. To analyze the effect of the diversity enhancement modules from three perspectives, we add patch-level weak-strong mixing augmentation(diverse-augment), diverse-scan collaborative Mamba (diverse-scan), and diverse-feature fusion module (diverse-feature), respectively. 
The results show the benefits of diversity enhancement from these three perspectives and they are significantly improved compared to the baseline (the first row of the table).

When we examine the impact of diverse-augment module alone, we observe a consistent improvement across all datasets. This indicates that introducing patch-level weak-strong mixing augmentation enhances the model's ability to handle variability in anatomical structures by exposing the model to more diverse patterns. On the Kvasir dataset, the slight increase in ASD (from 1.91 to 2.04) may suggest that some image transformations, while beneficial for improving segmentation performance, may have introduced minor boundary errors. Nevertheless, the overall segmentation quality still showed significant improvement.

The effect of diverse-scan, when applied independently, also demonstrates substantial improvements in segmentation performance. This module contributes by enabling the model to consider the spatial dependencies of organs from multiple scan directions. It could capture the cross-organ relationships for multi-organ Synapse datasets and help the model understand the spatial context and boundaries in single-object datasets.

Diverse-feature mechanism alone contributes to overall improvements in segmentation accuracy; however, it may introduce a certain level of complexity in the feature space, which could result in a slight increase in ASD. Nonetheless, when diverse-feature is combined with diverse-scan, the two networks foster feature differentiation across various scanning directions. It helps to refine the spatial relationships and improve segmentation boundaries.

The performance is markedly improved when all three components, diverse-augment, diverse-scan, and diverse-feature, are combined. For instance, on the Synapse dataset, the Dice score increases from 58.00\% to 67.63\%, while the ASD drops from 22.55 to 10.70, indicating a significant improvement in both segmentation accuracy and boundary precision.

\begin{table}[t]
\caption{Effectiveness of co-training in DCMamba on Synapse and ISIC18 datasets with 20\% labeled data.}
\label{tab: cotraining}
\centering
\scalebox{0.85}{
\begin{tabular}{c|cc|cc}
\hline
\noalign{\smallskip}
\multirow{2.5}{*}{Method} & \multicolumn{2}{c|}{Synapse} & \multicolumn{2}{c}{ISIC18} \\
\noalign{\smallskip}
& Dices $\uparrow$ & ASD $\downarrow$ & Dice $\uparrow$ & ASD $\downarrow$ \\ 
\noalign{\smallskip}
\hline
\noalign{\smallskip}
Different scans within a single network & 58.53 & 15.70 & 83.84 & 2.47 \\
\noalign{\smallskip}
\hline
\noalign{\smallskip}
Co-training networks with different scans & 61.62 & 19.81 & 84.78 & 1.87 \\
\noalign{\smallskip}
\hline
\noalign{\smallskip}
\textbf{DCMamba (Ours)} & \textbf{67.63} & \textbf{10.70} & \textbf{86.42} & \textbf{1.66} \\
\noalign{\smallskip}
\hline
\end{tabular}}
\end{table}

\begin{figure}[t]
\centering
\includegraphics[width = 0.48\textwidth]{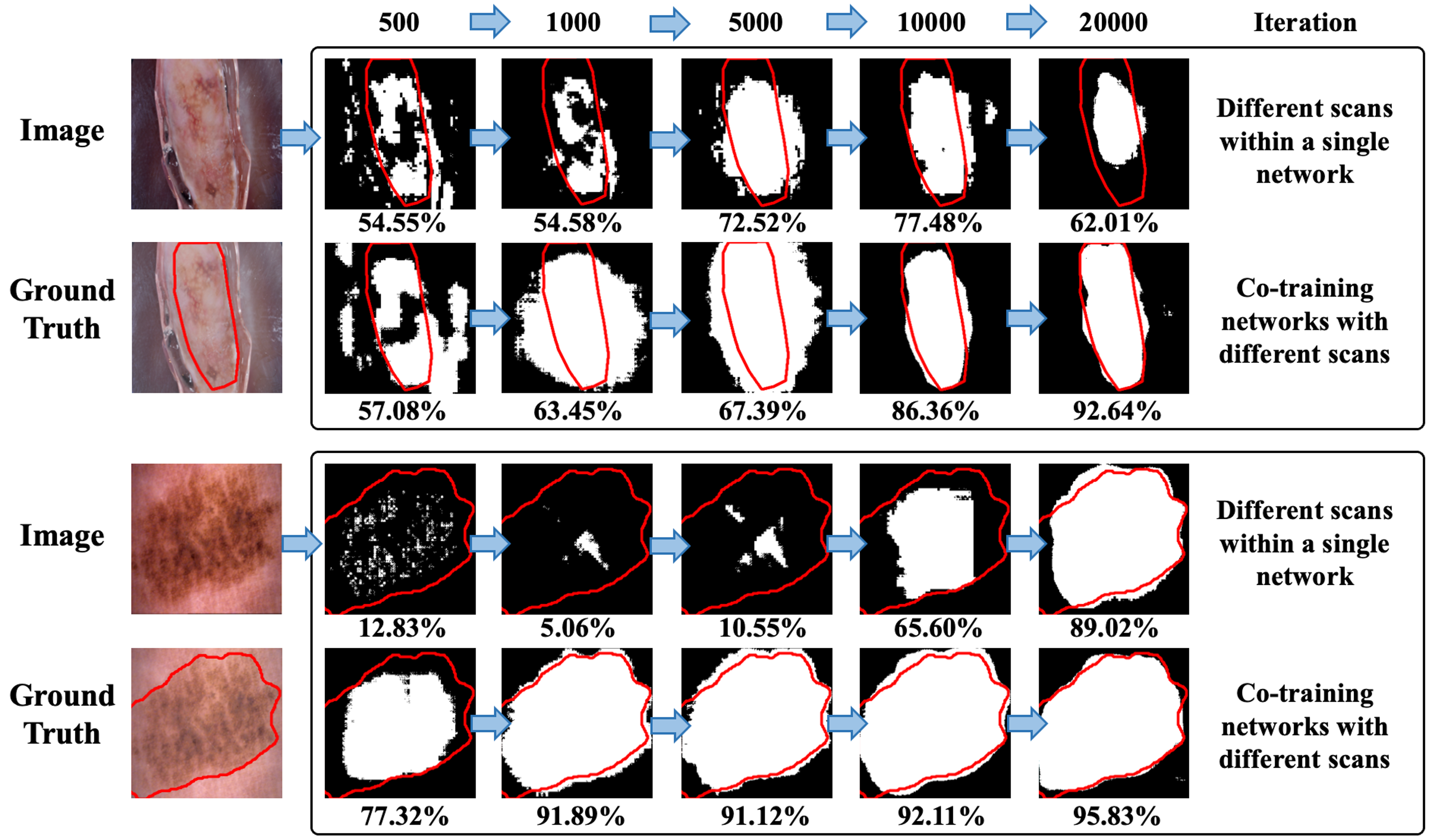}
\caption{Visualization of different scans within a single network and co-training networks.}
\label{fig: visual_cotraining}
\end{figure}

\begin{table}[t]
\caption{Effectiveness of diverse-feature contrastive fusion on ISIC18 dataset with 20\% labeled data.}
\label{tab: diverse_feat}
\centering
\scalebox{0.8}{
\begin{tabular}{c|cccc}
\hline
\noalign{\smallskip}
Method & mIoU (\%) $\uparrow$ & Dice (\%) $\uparrow$ & 95HD $\downarrow$ & ASD $\downarrow$ \\ 
\noalign{\smallskip}
\hline
\noalign{\smallskip}
contrastive loss only & 76.50 & 85.01 & 10.17 & 2.04 \\
contrastive loss with diverse-feature & \textbf{78.44} & \textbf{86.42} & \textbf{8.49} & \textbf{1.66} \\
\noalign{\smallskip}
\hline
\end{tabular}}
\end{table}

\begin{figure}[t]
   \centering
   \includegraphics[width = 0.48\textwidth]{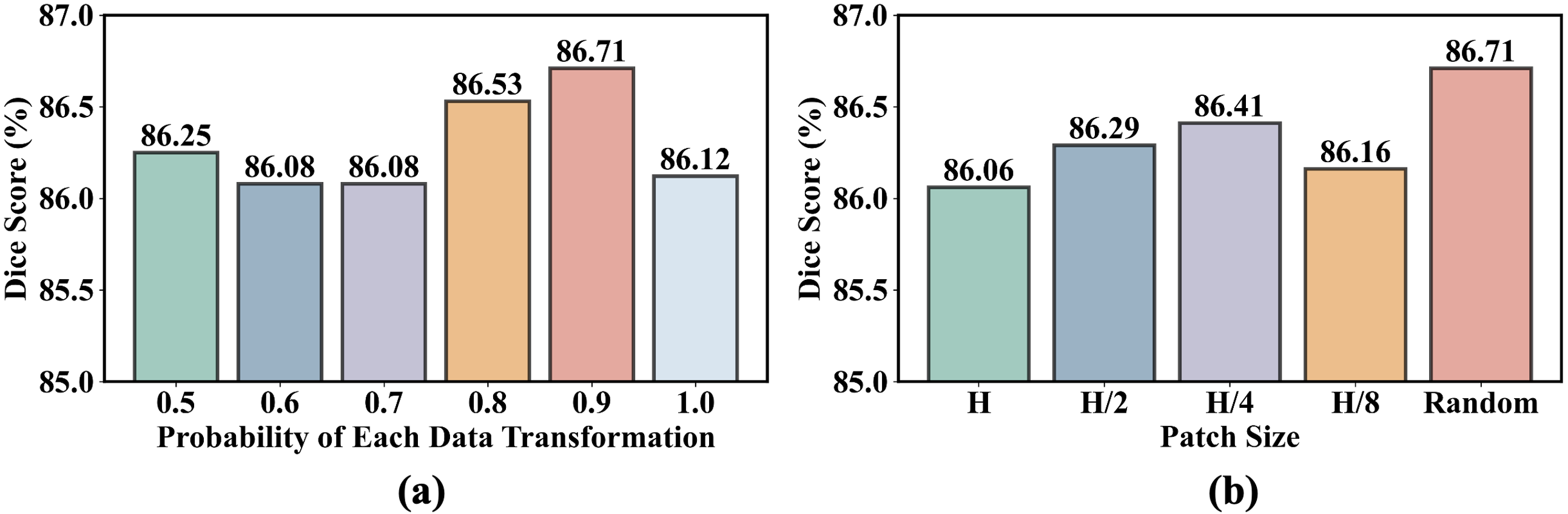}
   \caption{(a) The probability of each transformation of strong augmentation on ISIC18 dataset with 20\% labeled data. (b) Different patch sizes for patch-level augmentation. ``random" denotes randomly selecting patch sizes from $H/8$ to $H$ ($H=W$) in each iteration. }
   \label{fig: result_bar}
\end{figure}

\subsubsection{The Role of Collaborative Learning}
Our DCMamba framework employs a collaborative learning strategy between diverse scans to leverage complementary information from different image perspectives. To further explore the impact of collaborative learning between different scans, we compare it with the alternative approach of integrating different scans into a single network. The results, shown in Table~\ref{tab: cotraining}, clearly demonstrate the benefits of collaborative learning. On both the Synapse and ISIC18 datasets with 20\% labeled data, collaborative learning between different scans yields significant improvements. For example, the Dice score increases from 58.53\% to 67.63\% on the Synapse dataset, and from 83.84\% to 86.42\% on the ISIC18 dataset. Additionally, the ASD is reduced from 15.70 to 10.70 for Synapse, and from 2.47 to 1.66 for ISIC18, highlighting improved segmentation accuracy and boundary precision. The improvement could be attributed to the fact that collaborative learning allows the model to leverage complementary information from different scans independently, enhancing its ability to capture diverse features and handle complex spatial dependencies across different scan direction perspectives. In contrast, combining different scans into a single network may restrict the model's capacity to effectively integrate the unique characteristics of each scan, leading to suboptimal performance. As shown in Figure~\ref{fig: visual_cotraining}, the first case shows that the method of different scans in a single network gradually improves on vertical targets but eventually experiences performance degradation. The second case shows that the results of a single network are still insufficient in boundary details. Additionally, our diversity-enhanced strategy further strengthens this co-training approach by encouraging the network to capture varied and robust representations.

\begin{figure*}[t]
    \centering
    \includegraphics[width = 0.98\textwidth]{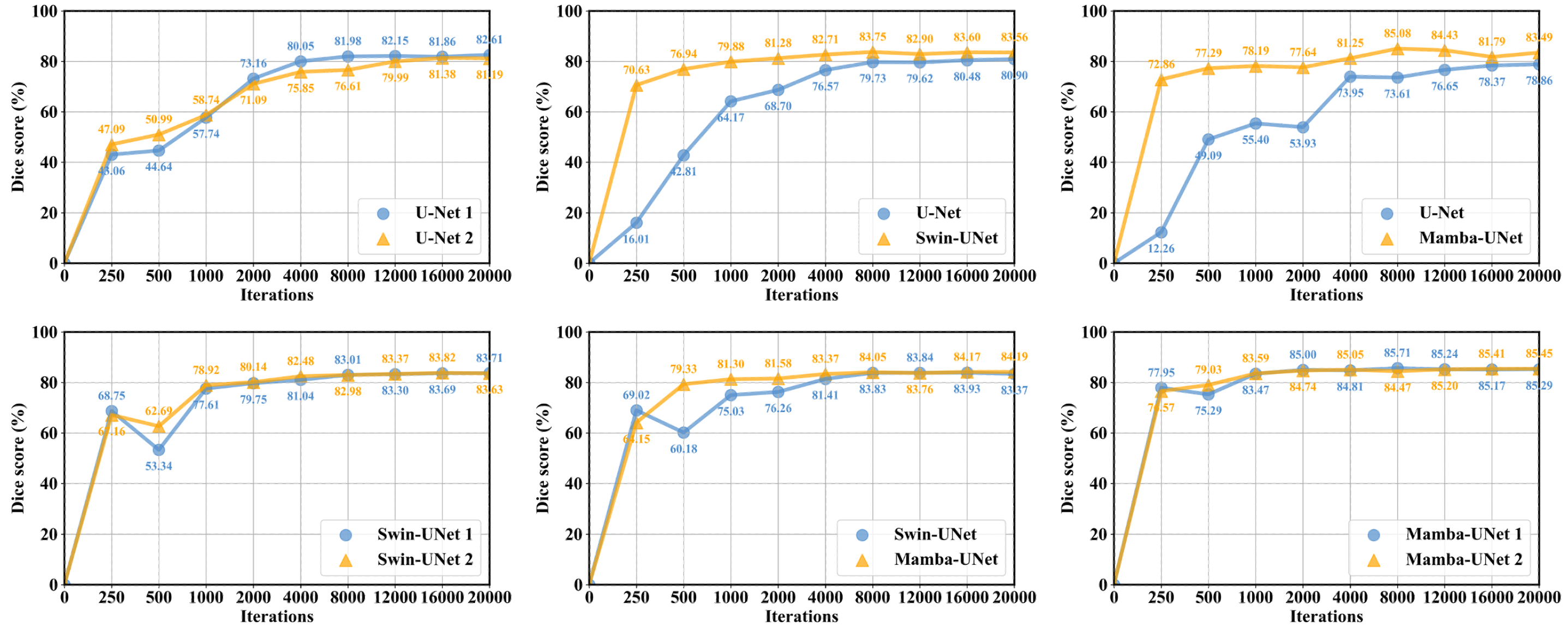}
    \caption{Collaborative learning across different backbones: CNN-based U-Net, Transformer-based Swin-UNet, and SSM-based Mamba-UNet.}
    \label{fig: diff_backbone}
\end{figure*}

\subsubsection{The Role of Uncertainty-weighted Contrastive Learning}
To verify the effectiveness of our uncertainty-weighted strategy, we compare the contrastive loss-only method and our uncertainty-weighted contrastive loss method in Table~\ref{tab: diverse_feat}. Our method increases the diversity between features and improves the Dice score from 85.01\% to 86.42\%.

\subsubsection{Probability of Strong Augmentation}
Our strong augmentations consist of pixel-wise transforms (Gaussian blur, brightness, contrast, gamma), with the probability of each transformation being applied to each sample represented by $\alpha$. We adjust $\alpha$ from 0.5 to 1.0 with a step of 0.1. The Dice scores of all experiments exceed 86\%, and the best results are obtained at 0.9, as shown in Figure~\ref{fig: result_bar}. 

\subsubsection{Different Patch Size}
We compare the diverse-augment strategy with different patch sizes in Figure~\ref{fig: result_bar}. The results indicate that randomly selecting patch sizes outperforms setting a fixed value. Random patch sizes enable the model to encounter augmented patches of various scales during training, enhancing its generalization capability.

\begin{table}[t]
\caption{Comparisons with other augmentation methods using different backbones evaluated by Dice score (\%).}
\label{tab: backbone}
\centering
\scalebox{0.85}{
\begin{tabular}{c|c|cc|cc}
\hline
\noalign{\smallskip}
\multirow{2.5}{*}{Method} & \multirow{2.5}{*}{Backbone} & \multicolumn{2}{c|}{20 / 80\%} & \multicolumn{2}{c}{10 / 90\%} \\ 
\noalign{\smallskip}
\cline{3-6}
\noalign{\smallskip}
& & Synapse & ISIC18 & Synapse & ISIC18 \\
\noalign{\smallskip}
\hline
\noalign{\smallskip}
\multirow{2}{*}{Fixmatch~\cite{sohn2020fixmatch}} & U-Net & 54.34 & 78.94 & 50.22 & 76.97 \\
& Mamba-UNet & 63.50 & 84.71 & 59.27 & 84.01 \\
\noalign{\smallskip}
\hline
\noalign{\smallskip}
\multirow{2}{*}{BCP~\cite{bai2023bidirectional}} & U-Net & 57.92 & 84.41 & 55.26 & 83.13 \\
& Mamba-UNet & 60.23 & 84.58 & 54.69 & 82.45 \\
\noalign{\smallskip}
\hline
\noalign{\smallskip}
\multirow{2}{*}{CML~\cite{wu2024cross}} & U-Net & 47.64 & 81.54 & 40.00 & 83.08 \\
& Mamba-UNet & 59.39 & 83.75 & 50.39 & 82.97 \\
\noalign{\smallskip}
\hline
\noalign{\smallskip}
\cellcolor{gray!20}\textbf{DCMamba (Ours)} & \cellcolor{gray!20}\textbf{Mamba-UNet} & \cellcolor{gray!20}\textbf{67.63} & \cellcolor{gray!20}\textbf{86.71} & \cellcolor{gray!20}\textbf{61.62} & \cellcolor{gray!20}\textbf{85.31} \\
\noalign{\smallskip}
\hline
\end{tabular}}
\end{table}

\begin{table}[t]
\caption{Comparison with foundation models evaluated\\ by Dice score (\%).}
\label{tab: foundation}
\centering
\scalebox{0.9}{
\begin{tabular}{c|c|c|c|c|c}
\hline
\noalign{\smallskip}
Method & Prompt & Synapse & ISIC17 & ISIC18 & Kvasir \\
\noalign{\smallskip}
\hline
\noalign{\smallskip}
\multirow{3}{*}{SAM~\cite{kirillov2023segment}} & 1 point & 30.97 & 36.79 & 38.81 & 37.40 \\
& 5 points & 39.27 & 40.83 & 45.98 & 44.38 \\ 
& 10 points & 27.74 & 40.02 & 48.83 & 45.09 \\ 
\noalign{\smallskip}
\hline
\noalign{\smallskip}
\multirow{3}{*}{SAM-Med2D~\cite{cheng2023sam}} & 1 point & 66.33 & 42.51 & 43.86 & 41.43 \\
& 5 points & 64.66 & 50.90 & 52.41 & 47.42 \\
& 10 points & \textbf{67.84} & 57.30 & 61.78 & 54.23 \\
\noalign{\smallskip}
\hline
\noalign{\smallskip}
\multicolumn{2}{c|}{SemiSAM~\cite{zhang2023semisam} - 10\% labeled} & - & 66.65 & 76.71 & 61.74 \\ 
\multicolumn{2}{c|}{DCMamba (Ours) - 10\% labeled} & 61.62 & 79.48 & 83.94 & 82.83 \\ 
\noalign{\smallskip}
\hline
\noalign{\smallskip}
\multicolumn{2}{c|}{SemiSAM~\cite{zhang2023semisam} - 20\% labeled} & - & 72.60 & 80.32 & 75.69 \\ 
\multicolumn{2}{c|}{DCMamba (Ours) - 20\% labeled} & 67.63 & \textbf{81.05} & \textbf{86.42} & \textbf{86.44} \\ 
\noalign{\smallskip}
\hline
\end{tabular}}
\end{table}

\subsection{Further Analysis}
\subsubsection{Comparisons with SSM-based SSL Methods}
To further analyze our framework in semi-supervised segmentation, we compare it with three SSL methods involving strong augmentation, ensuring they have the same backbone. Specifically, we replace the backbone of Fixmatch~\cite{sohn2020fixmatch}, BCP~\cite{bai2023bidirectional}, and CML~\cite{wu2024cross} with the SSM-based Mamba-UNet~\cite{wang2024mamba}, as shown in Table~\ref{tab: backbone}. This ensures that the comparison is fair and that the performance improvements are due to the methods themselves, rather than the backbone architecture.

First, the results show that the Mamba-UNet backbone consistently outperforms the standard U-Net backbone across all methods and datasets. In addition, with the effectiveness of our diversity enhancement strategies, our DCMamba outperforms all the other augmentation methods with a U-Net backbone and SSM-based backbone. Also, DCMamba maintains a relatively strong performance as the reduced amount of labeled data (from 20\% to 10\%), which indicates the robustness of our approach in semi-supervised settings.

\subsubsection{Comparisons with Foundation Models}
We compare our method with foundation models, SAM~\cite{kirillov2023segment} and SAM-Med2D~\cite{cheng2023sam}. SAM-Med2D fine-tunes SAM on medical datasets, including the four datasets (Synapse, ISIC17, ISIC18, Kvasir) used in our experiments. As illustrated in Table~\ref{tab: foundation}, our DCMamba exhibits slightly lower performance than SAM-Med2D on the Synapse dataset, while it outperforms foundation models with 1, 5, or 10-point prompts on the other three datasets with 20\% annotations. 

In addition, we also conduct the comparison with SAM-based SSL methods SemiSAM~\cite{zhang2023semisam}. The U-Net~\cite{ronneberger2015u} is used as the semi-supervised backbone and SAM-Med2D~\cite{cheng2023sam} is as the SAM backbone for 2D segmentation. This method is designed for single-object segmentation and we evaluated the performance of the ISIC17, ISIC18, and Kvasir datasets. It could be observed that our method outperforms SemiSAM, achieving superior segmentation accuracy.

\subsubsection{Collaborative Learning Across Different Backbones}
To further investigate the benefits of collaborative learning for Mamba network in semi-supervised medical image segmentation, we conduct co-training experiments comparing it with other backbones, including CNN-based and Transformer-based networks. The results are presented in Figure~\ref{fig: diff_backbone}, and it could be found that when conducting co-training with networks using CNN (U-Net) and Transformer/SSM-based (Swin-UNet/Mamba-UNet) networks, the CNN-based branch rises slowly and performs worse than the Transformer/SSM-based branch. This can be attributed to the effect of pre-training, where the Transformer/SSM-based models are pre-trained on large natural image datasets and benefit from initialization. Furthermore, when combining Swin-UNet with Mamba-UNet networks in collaborative learning, we found that the performance and stability of Swin-UNet slightly lag behind that of Mamba-UNet. The promising co-training performance is achieved when both branches utilize the SSM-based backbone. This also indicates the promising synergy between Mamba and co-training in semi-supervised tasks.

\subsubsection{Diversity in Co-training}
Previous studies~\cite{wang2007analyzing} have shown that co-training can remain effective even when the two classifiers have large diversity. Recent works~\cite{luo2022semi, wang2022cnn} further demonstrate that using different model architectures (\eg, CNN and Transformer, or CNN and Mamba) in co-training improves performance due to their diverse inductive biases. In our case, although the diverse-scan dual-branch design introduces differences by applying different scan directions, the resulting diversity is more subtle than using entirely different backbones. This level of diversity helps reduce confirmation bias between branches without destabilizing training.

\section{Conclusion}
In this paper, we propose DCMamba, a novel diversity-enhanced semi-supervised medical image segmentation framework. 
DCMamba effectively increases diversity from data, network, and feature perspectives. Specifically, we introduce patch-level weak-strong mixing augmentation to enrich data diversity, a diverse-scan collaborative Mamba module to capture spatial information through multiple scanning directions, and a diverse-feature fusion strategy to effectively integrate heterogeneous features.
It fully capitalizes on the cross-supervision potential of the two SSM-based branches. 
Experimental results across multiple datasets demonstrate that our method achieves significant performance improvements in both 2D and 3D medical image segmentation tasks.


\bibliographystyle{IEEEtran}
\bibliography{tmi}

\end{document}